\documentclass[11pt]{article}

\usepackage[preprint]{acl}
\usepackage{times}
\usepackage{latexsym}
\usepackage[T1]{fontenc}
\usepackage[utf8]{inputenc}
\usepackage{microtype}
\usepackage{amsmath}
\usepackage{booktabs}
\usepackage{multirow}
\usepackage{graphicx}
\usepackage{xcolor}
\usepackage{soul}
\usepackage{enumitem}
\usepackage{stfloats}
\usepackage{etoc}

\newcommand{\jev}{Jev}
\newcommand{\luna}{Luna}

\definecolor{findbg}{HTML}{EEF4FC}
\definecolor{findrule}{HTML}{9DBDE6}
\newcounter{finding}

\usepackage{tcolorbox}
\tcbuselibrary{skins,breakable}
\definecolor{pcgrey}{HTML}{8C949E}
\definecolor{pcforced}{HTML}{5F97D2}
\definecolor{pcjev}{HTML}{D76364}
\definecolor{pcslot}{HTML}{6B6CC0}
\definecolor{pcback}{HTML}{F5F7FA}
\providecommand{\linenumbers}{}\providecommand{\nolinenumbers}{}
\newtcolorbox{pcbox}[2]{enhanced,breakable,frame hidden,colback=pcback,sharp corners,boxrule=0pt,
  borderline west={2pt}{0pt}{#1},left=6pt,right=5pt,top=3pt,bottom=4pt,before skip=5pt,after skip=7pt,
  fontupper=\fontfamily{zi4}\footnotesize\selectfont\raggedright\frenchspacing\microtypesetup{protrusion=false},
  before upper={{\normalfont\sffamily\footnotesize\bfseries\color{#1!75!black}#2}\par\vspace{2pt}}}
\newenvironment{promptcard}[2]{\par\nolinenumbers\begin{pcbox}{#1}{#2}}{\end{pcbox}\par\linenumbers}
\newcommand{\pcpart}[1]{\par\vspace{3pt}{\normalfont\sffamily\scriptsize\color{black!55}#1}\par\vspace{1pt}}
\newcommand{\pslot}[1]{{\color{pcslot}\{#1\}}}
\newcommand{\pgap}{\par\vspace{0.5\baselineskip}}
\newcommand{\pstub}[1]{\textbf{#1}}
\newcommand{\pnl}{{\color{black!45}\textbackslash n}}
\newcommand{\pcnote}[1]{\par\vspace{3pt}{\normalfont\sffamily\scriptsize\color{black!55}#1}\par}

\title{Truncated Reasoning Under a Budget: \\Evaluating System-1 Models as Decision-Makers}

\author{
  \textbf{Zhongyin Wang\textsuperscript{1}\thanks{
    Corresponding authors: Zhongyin Wang and Lingyao Li.
    Emails: zw92@illinois.edu, lingyaoli@arizona.edu.
  }},
  \textbf{Zhaoqian Xue\textsuperscript{2}},
  \textbf{Junting Mu\textsuperscript{3}},
  \textbf{Haozheng Liu\textsuperscript{4}},
\\
  \textbf{Ye Yuan\textsuperscript{5,6}},
  \textbf{Haolun Wu\textsuperscript{5,6}},
  \textbf{Lingyao Li\textsuperscript{2}\footnotemark[2]}
\\[0.5ex]
  {\normalfont\normalsize
  \textsuperscript{1}University of Illinois Urbana-Champaign,
  \textsuperscript{2}University of Arizona}
\\
  {\normalfont\normalsize
  \textsuperscript{3}University of California, Irvine,
  \textsuperscript{4}New York University}
\\
  {\normalfont\normalsize
  \textsuperscript{5}Mila -- Quebec Artificial Intelligence Institute,
  \textsuperscript{6}McGill University}
}

\begin{document}
\maketitle

\begin{abstract}
Large language models (LLMs) often generate more reasoning tokens than needed. Limiting the reasoning budget reduces cost, but forcing an answer from an incomplete trace can hurt accuracy. System-1 decision models select among candidate options without producing a reasoning response. We evaluate \jev{} as a decision reader of truncated reasoning across seven LLM reasoners and five multiple-choice benchmarks (35 pairs). \jev{} achieves higher mean accuracy than forced answering at tested budgets from 500 to 16,000 tokens, including the estimated costs of reasoning and answering. At a 1{,}500-token budget, it improves accuracy by 4.1~pp overall and 8.6~pp on truncated questions, with significant gains in 17 pairs under Holm-corrected bootstrap tests. Offline replay estimates 34.5\% of the full-run cost, with a 1.13~pp accuracy loss. A policy selected on training folds saves 71.3\% of the estimated full-run cost on held-out folds, with a 0.58~pp accuracy loss. We also find that \jev{} shows strong question-only accuracy, exceeding 500-token forced answering on average (68.0\% vs.\ 65.5\%), although this advantage varies by benchmark. These findings suggest that System-1 decision models can complement LLM reasoning and reduce the need for costly reasoning completion.
\end{abstract}

\section{Introduction}
\label{sec:intro}
\vspace{-1mm}

\begin{figure}[t]
\centering
\includegraphics[width=\columnwidth]{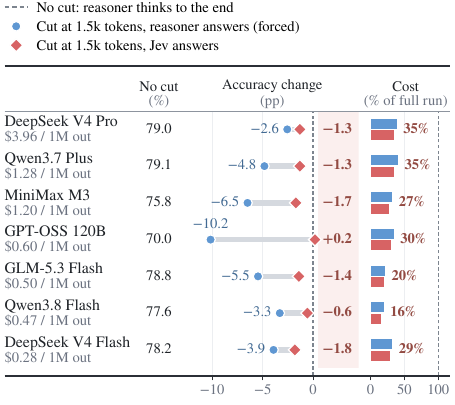}
\caption{Seven LLM reasoners cut at 1{,}500 tokens (mean over five benchmarks). Left: full-run accuracy and the change when the reasoner answers (blue) or \jev{} answers (red). Right: cost as \% of the uncapped run.}
\vspace{-6mm}
\label{fig:teaser}
\end{figure}

Large language models (LLMs) can now reason before they answer. Reasoning models such as o1 and DeepSeek-R1 generate extended chains of thought to tackle complex problems~\citep{openai2024reasoning,deepseek2025r1}. Some models also expose adjustable thinking budgets to balance performance and latency~\citep{qwen2025qwen3}. However, models may generate redundant reasoning that adds little to answer quality while consuming additional tokens, a phenomenon
known as overthinking~\citep{chen2025overthinking}. Token-budget prompting~\citep{han2025tale},
budget forcing~\citep{muennighoff2025s1}, and early-exit rules~\citep{liu2025answer,yang2026dynamic,fu2025deepconf,sun2026stop}
seek to reduce this cost by controlling how much reasoning is generated. Yet once the reasoning stops, the reasoner must still answer from an unfinished trace, and our results show that the answering mechanism substantially affects accuracy at short budgets.

Recently, System-1 decision models such as \jev{} \citep{typesafe2026jev} and the GPT-6 Luna Decision \citep{openai2026decisions} have offered a different approach. They score a fixed set of options in one forward pass without generating text, at a fraction of an LLM's cost. \jev{} has been evaluated as a stand-alone decision model across classification, reading comprehension, and reasoning tasks
\citep{deusser2026jev,duan2026general}.
It has also been studied as a judge~\citep{li2026jevjudge}
and as a decision layer in LLM agents~\citep{wu2026reflex}. These tasks involve choosing among known options using information already in the input, such as grading a response or selecting the next action, making them well suited to fast, low-cost decision models.

How decision models use partially generated reasoning and whether this improves cost--accuracy trade-offs remains less understood. An LLM reasoner produces evidence at a high cost, whereas a decision model can read that evidence at a much lower cost~\citep{kahneman2011thinking}. Prior work shows that reasoning traces can be reused by other models~\citep{cheng2026travels} and that decision models are most competitive when decisions can be resolved from available
evidence~\citep{duan2026general}.
These findings motivate a division of labor in which a reasoner generates intermediate text and a decision model selects the final answer, and we ask three research questions:
\begin{itemize}[leftmargin=*, nosep]
    \item \textbf{RQ1 (Reliability):}
    How reliably can a decision model answer from truncated reasoning compared with the reasoner and other readers?
    \item \textbf{RQ2 (Cost--accuracy trade-off):}
    What cost--accuracy trade-offs arise under static budgets, and what benefit does adaptive stopping add?
    \item \textbf{RQ3 (Trace contribution):}
     When and to what extent does the reasoning trace improve or degrade the decision model's answers?
\end{itemize}

We study these questions in multiple-choice question answering across seven LLM reasoners and five benchmarks \citep{rein2024gpqa,du2025supergpqa,jin2021medqa,zuo2025medxpertqa,wang2024mmlupro}. We stop each reasoner after a fixed token budget and use a decision model to select an answer based on the truncated trace. We compare this approach with forcing the LLM reasoner to answer. For each approach, we include reasoning and answering costs. Across all seven reasoners, \jev{} improves accuracy over forced answering. At 1{,}500-token budget, it achieves 75.8\% accuracy, compared with 71.7\% for forced answering, at about one-third the cost of uncapped reasoning (\S\ref{sec:same}, \S\ref{sec:cost}). We also find that \jev{} can answer many questions correctly without a trace, while longer reasoning remains beneficial on the hardest benchmark (\S\ref{sec:prior}). These results suggest that pairing an expensive reasoner with a fast decision model has the potential to reduce reasoning costs with little loss in performance.

\section{Related Work}
\label{sec:related}
\vspace{-1mm}

\paragraph{LLM Overthinking and Reasoning Efficiency}

Long reasoning traces can improve LLMs' problem solving, but also cause overthinking, increasing computation and latency \citep{chen2025overthinking}. Existing work allocates computation before reasoning through model routing \citep{ong2025routellm}, reasoning mode selection \citep{fang2025thinkless,lou2025adacot}, and token budget control \citep{han2025tale,muennighoff2025s1}. However, these allocations may not match the needs that emerge during reasoning.

Dynamic stopping methods instead monitor answer the stability \citep{liu2025answer,sun2026stop}, confidence or prediction signals \citep{kim2026think,yang2026dynamic}, or learn stopping policies \citep{dong2026does}. Methods requiring hidden states or token probabilities are difficult to apply when APIs expose only reasoning text. The black-box variant of EAT \citep{wang2025entropy} uses a smaller proxy model to compute entropy-based stopping signals while retaining the original reasoner for final answering. This enables external stopping decisions, while leaving answerer choice as a separate design consideration.

\par\vspace{2mm}
\noindent\textbf{System-1 Decision Models}
System-1 decision models directly score candidate answers without autoregressively generating responses, potentially reducing cost and latency. \citet{deusser2026jev} benchmark Jev across classification, reading comprehension, and common sense reasoning. \citet{duan2026general} find that decision models are competitive when available evidence supports a decision, but remain limited on tasks requiring specialized knowledge. This motivates separating standalone answering ability from the contribution of supplied reasoning. Prior work also explores model collaboration. Jev-Mem \citep{jiang2026jev} uses System-1 to control memory retrieval and System-2 for reasoning and answer generation. Reasoning that Travels \citep{cheng2026travels} studies prefix transfer between generative models and receiver agreement for early stopping. We compare non-generative \jev{} with the original reasoner as final answerers at matched prefixes, evaluating accuracy, monetary cost, and latency, with question-only as a control for the contribution of external reasoning.
\begin{figure*}[t]
\centering
\includegraphics[width=\textwidth]{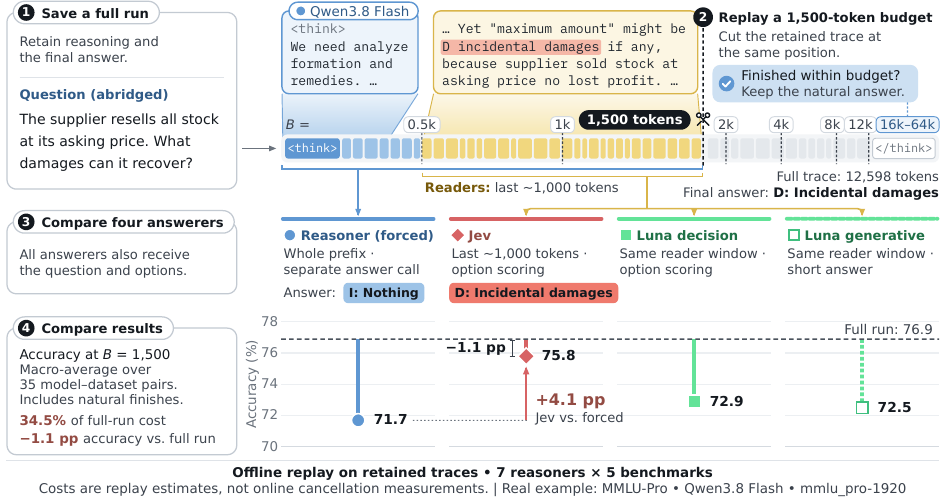}
\caption{
Comparing four answerers by offline replay at a 1{,}500-token budget.
Runs with a valid natural answer within the budget retain that answer for every policy. At a cut, the original reasoner receives the whole prefix, while readers receive the question, options, and the final approximately 1{,}000 reasoning tokens.}
\vspace{-4mm}
\label{fig:framework}
\end{figure*}

\section{Methods and Experiments}
\label{sec:method}
\vspace{-1mm}

\subsection{Problem Setting}
Given a question $q$ and options $\mathcal O$, LLM reasoner $M$ produces a retained reasoning trace $z_{1:L}$ and, when available, a final answer $a_{\mathrm{full}}$.
We compare answerers on prefixes of this trace without accessing model weights or hidden states.
The reasoning budget $B$, checkpoint interval $k$, and reader window $w$ control generation length, checking frequency, and reader context, respectively.
Figure~\ref{fig:framework} summarizes the comparison.

\subsection{Benchmarks}
Our benchmark consists of five datasets: GPQA-Diamond (198 questions; \citealp{rein2024gpqa}), SuperGPQA (500; \citealp{du2025supergpqa}), MedQA (500; \citealp{jin2021medqa}), MedXpertQA-Text (500; \citealp{zuo2025medxpertqa}), and MMLU-Pro (500; \citealp{wang2024mmlupro}).
SuperGPQA contains 167 easy, 166 middle, and 167 hard questions, reweighted to the full benchmark's difficulty distribution.
Together with seven reasoners, these yield 35 reasoner--benchmark pairs.

The retained acquisition cohort contains 15{,}377 model--question runs, including 15{,}309 with a valid natural answer. The other 68 full-run outcomes, including one saved generation failure, count as incorrect. The four-way comparison at $B=1.5$k uses 15{,}376 matched runs: one outcome lacks an exact forced answer and is excluded from this comparison, not from the acquisition cohort. Trace acquisition and missing-data accounting appear in Appendix~\ref{app:data}.

\subsection{Model Selection}
\label{sec:answer}
\paragraph{LLM as Reasoners.}
We use DeepSeek V4 Pro \citep{deepseek2026v4pro0813},
Qwen3.7 Plus \citep{qwen37plus}, MiniMax M3 \citep{minimax2026m3},
GPT-OSS 120B with high reasoning effort \citep{openai2025gptoss},
GLM-5.3 Flash \citep{zai2026glm53flash},
Qwen3.8 Flash \citep{alibaba2026qwen38flash}, and
DeepSeek V4 Flash \citep{deepseek2026v4flash0731},
all accessed through Together AI \citep{together2026availablemodels}.
Forced answering supplies the whole prefix to the reasoner in a separate request, using a provider-supported strict or think-close protocol (Appendix~\ref{app:prompts}).

\par\vspace{2mm}
\noindent\textbf{Decision and Generative Readers.}
We evaluate three reader configurations: \jev{} \citep{typesafe2026jev}, a non-generative System-1 decision model, and GPT-6 Luna accessed through either its Decisions API \citep{openai2026decisions} or generative interface \citep{openai2026luna}. \jev{} supports 64k input tokens per request, with a 32k limit on the state plus the longest decision question.
All three receive the question, options and the final nominal $w=1{,}000$ reasoning tokens before the cut. The decision interfaces return option choices. Luna's generative interface returns a short answer with confidence. Reader comparisons match the saved reasoning window, where the main forced-answer comparison also changes context length and answer elicitation.

\subsection{Experiment Setup and Cost Estimation}
\label{sec:replay}
\label{sec:policies}
\paragraph{Cost estimation and experiment procedure.}
A fixed budget $B$ retains $a_{\mathrm{full}}$ only when a valid natural answer is available and $L\le B$.
Otherwise, answerers use the cut $c=\min(B,L)$, except that a saved generation failure reached before answering is scored incorrect without inventing an additional call.
At $B=1.5$k, forced answers matched the exact saved cut. One outcome with an unavailable exact forced answer is omitted from the four-way comparison. Other positive budgets retain the nearest earlier forced answer when an exact record is unavailable (Appendix~\ref{app:exact}).
Relative-progress diagnostics use $c=\operatorname{round}(fL)$ for $f\in\{0.1,\ldots,1.0\}$ and invoke an answerer even at $f=1$. Here $L$ is the observed saved trace length, including capped or failed traces, not necessarily a natural completion. These cuts are retrospective diagnostics.
Table~\ref{tab:replay-procedure} gives the replay procedure. Answerer inputs exclude later reasoning, separately generated final-answer content and gold labels, although the retained prefix may itself contain an explicit answer.

\begin{table}[t]
\centering
\small
\begin{tabular}{@{}r@{\quad}p{0.87\columnwidth}@{}}
\toprule
\multicolumn{2}{@{}l@{}}{\textbf{Experiment procedure}} \\
\midrule
& \textbf{Input:} retained traces, answer records, cut settings, window $w$. \\
1 & \textbf{for each} trace and budget $B$ or fraction $f$ \textbf{do} \\
2 & \quad $c\gets\min(B,L)$ or $\operatorname{round}(fL)$. \\
3 & \quad \textbf{if} budget mode and a saved generation failure is reached before answering \textbf{then} \\
4 & \qquad Record failure. Do not invent an answer call. \\
5 & \quad \textbf{else if} budget mode, $L\le B$, and a valid natural answer exists \textbf{then} \\
6 & \qquad Retain that answer for all answerers. \\
7 & \quad \textbf{else} \\
8 & \qquad Retrieve the forced answer at $c$. Require an exact cut at $B=1.5$k, otherwise allow the earlier-position fallback. \\
9 & \qquad Omit the outcome from the four-way 1.5k comparison if its exact forced answer is missing. \\
10 & \qquad Retrieve each reader's answer from $(q,\mathcal O,\operatorname{Tail}_{w}(z_{\le c}))$. \\
11 & \quad \textbf{end if}\\ 
12 & \quad Record correctness and each policy's cost. \\
13 & \textbf{end for} \\
\bottomrule
\end{tabular}
\caption{Offline replay using precollected answers. Policy cost includes only calls used by that policy.}
\vspace{-4mm}
\label{tab:replay-procedure}
\end{table}

\par\vspace{2mm}
\noindent\textbf{Adaptive stopping.} 
Adaptive policies query \jev{} every $k=500$ tokens through 16k, strictly before $\min(B,L)$.
Stopping uses confidence and answer stability, settled/useful scores, or their logical OR (Appendix~\ref{app:replayrules}).
A trigger submits \jev{}'s current choice directly. Otherwise, the budget rule applies.
All checks are charged, including those preceding natural finishes.

\par\vspace{2mm}
\noindent\textbf{Cost and latency.}
A policy specifies the reasoning budget, any intermediate
stopping checks, and the final answerer. The policy cost is calculated as:
\begin{equation}
C=C_{\mathrm{reasoner}}+\sum_{r\in\mathcal R}C_{\mathrm{reader}}(r)+C_{\mathrm{forced}},
\vspace{-2mm}
\label{eq:cost}
\end{equation}
where $\mathcal R$ contains the policy's reader checks and final calls.
Costs cover the original prompt, retained output, and billable answerer usage, including caching and retries. $C_{\mathrm{forced}}=0$ when a reader answers.
Offline acquisition expenses are excluded.
Fixed-budget latency is estimated from retained reasoning tokens, measured decoding speed, and answer-call latency.
These replay estimates do not establish online cancellation savings or asynchronous checking overhead.

\subsection{Evaluation Protocol and Metrics}
We report overall and cut-only accuracy, accuracy differences in percentage points (pp), and cost savings $100(1-\overline C_{\pi}/\overline C_{\mathrm{full}})$ within each pair. We also report estimated fixed-budget latency in seconds and as a percentage of the full run. Cross-pair summaries weight the 35 pairs equally.

\label{sec:selection}
For tuned caps and adaptive policies, we use five-fold cross-validation within each pair. In each split, four training folds select the lowest-cost policy subject to at most 1pp accuracy loss relative to the full run. If no candidate is feasible, the candidate with the highest training accuracy is selected.
The selected policy is evaluated on the remaining held-out fold, and each question's outcomes are averaged across the five repeats. The training constraint does not guarantee the same bound on held-out folds. Search grids appear in Appendix~\ref{app:methoddetails}.

In addition, paired intervals use 2{,}000 question-bootstrap resamples, stratified and weighted by difficulty on SuperGPQA. Positive-budget answerer comparisons use 100{,}000 resamples. The $B=0$ row retains the earlier question-only analysis. For the primary \jev{}--forced comparison at $B=1.5$k, we apply Holm's correction across the 35 pairs \citep{holm1979simple}. Other intervals are uncorrected (Appendix~\ref{app:multiple}). Trace contribution is measured against each reader's question-only answer on matched questions. Net gain equals the fix rate minus the hurt rate under a common denominator.

\section{Results}
\label{sec:results}
\vspace{-1mm}

\subsection{RQ1: Answer from truncated reasoning}
\label{sec:same}

\begin{figure*}[t]
\centering
\includegraphics[width=\textwidth]{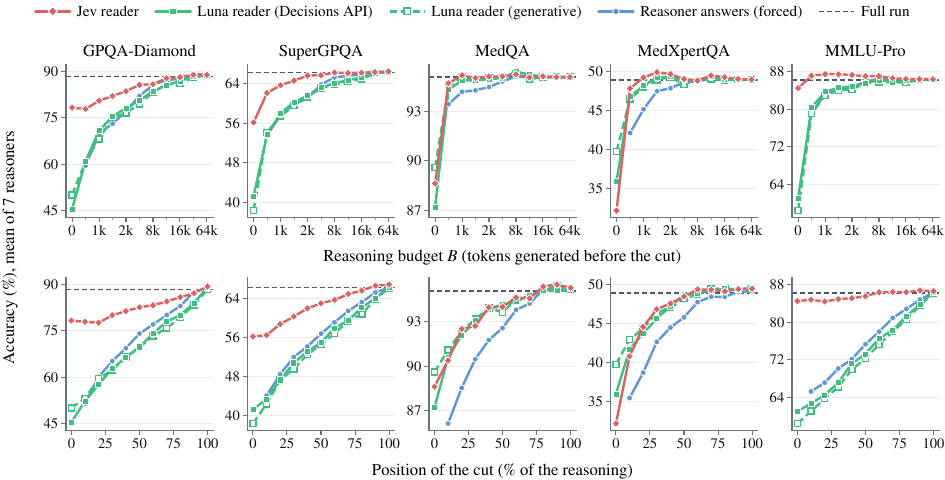}
\caption{Accuracy of the four answerers, mean of the seven reasoners per benchmark.
Top: fixed budget $B$, naturally completed runs keep the reasoner's answer, so the curves approach one another at large budgets. Bottom: diagnostic cuts at a fraction of each saved trace's length, known only after generation, and readers are evaluated even at 100\%, rather than replaced by the natural answer. Readers see the last approximately 1{,}000 tokens, where forced answers generally use the whole prefix. Dashed line: full-run accuracy. Per-pair panels are in Appendix~\ref{app:grid}.}
\vspace{-4mm}
\label{fig:answerers}
\end{figure*}

\begin{table*}[t]
\centering\small
\setlength{\tabcolsep}{5pt}
\begin{tabular}{r cccc cc ccc}
\toprule
 & \multicolumn{4}{c}{Accuracy (\%)} & \multicolumn{2}{c}{Cost (\% of full run)} &
 \multicolumn{3}{c}{\jev{} significantly better / worse (of 35 pairs)} \\
\cmidrule(lr){2-5}\cmidrule(lr){6-7}\cmidrule(lr){8-10}
$B$ & \jev{} & \luna{}-dec. & \luna{}-gen. & Forced & \jev{} & Forced & vs.\ forced & vs.\ \luna{}-dec. & vs.\ \luna{}-gen. \\
\midrule
0 & \textbf{68.0} & 54.2 & 55.2 & 53.4 & 1.7 & 4.7 & 28 / 1 & 21 / 2 & 21 / 7 \\
0.5k  & \textbf{73.9} & 67.2 & 67.0 & 65.5 & 19.1 & 22.2  & 29 / 0  & 22 / 0  & 22 / 0  \\
1k    & \textbf{75.2} & 71.1 & 70.3 & 70.1 & 28.2 & 31.4  & 25 / 0  & 20 / 0  & 21 / 0  \\
1.5k  & \textbf{75.8} & 72.9 & 72.5 & 71.7 & 34.5 & 38.3  & 21 / 0  & 14 / 1  & 20 / 0 \\
2k    & \textbf{76.3} & 73.8 & 73.3 & 73.1 & 39.5 & 43.4  & 18 / 0  & 16 / 0  & 18 / 0  \\
4k    & \textbf{76.5} & 74.7 & 74.3 & 75.0 & 53.2 & 57.7  & 10 / 0  & 13 / 0  & 15 / 0  \\
8k    & \textbf{76.6} & 75.6 & 75.4 & 76.2 & 68.8 & 73.8  & 3 / 0   & 7 / 0   & 11 / 1  \\
16k   & \textbf{77.0} & 76.5 & 76.1 & 76.8 & 84.1 & 88.3  & 4 / 0   & 6 / 0   & 6 / 0   \\
32k   & 77.1 & 77.0 & 76.9 & 77.2 & 94.4 & 96.4  & 0 / 1   & 0 / 0   & 0 / 0   \\
64k   & 77.2 & 77.1 & 77.1 & 77.1 & 99.8 & 101.3 & 0 / 0   & 0 / 0   & 1 / 0   \\
\bottomrule
\end{tabular}
\caption{Fixed-budget policies, macro-average over 35 reasoner--benchmark pairs (full run 76.9\%).
Each policy's estimated cost includes its retained generation and answering calls. Cost percentages are averaged within pairs, not ratios of pooled dollar totals. Significance counts refer to pairs whose uncorrected 95\% paired-bootstrap interval for \jev{}-minus-comparator accuracy lies entirely above or below zero.}
\vspace{-4mm}
\label{tab:main}
\end{table*}

Unless noted otherwise, results are macro-averages over the 35 pairs on the retained cohort of 15{,}377 model--question runs. Fixed-budget policies retain natural answers, whereas the percentage view evaluates truncated reasoning for every question. These are distinct evaluations, not interchangeable ways of setting an online budget.

\par\vspace{2mm}
\noindent\textbf{Answering at a fixed cut.}
\jev{} has the highest mean accuracy among the four answerers at every tested budget from 0.5k to 16k tokens (Table~\ref{tab:main}). Its advantage over forced answering decreases from 8.4pp at 0.5k tokens to 5.1pp at 1k, 4.1pp at 1.5k and 3.2pp at 2k. At 1.5k, the respective accuracies are 75.8\% and 71.7\%, compared with 72.9\% for \luna{} decision and 72.5\% for \luna{} generative.
Under the uncorrected 95\% paired-bootstrap interval, \jev{} has higher accuracy than forced answering in 21 of 35 pairs and lower accuracy in none. Averaging over benchmarks within each reasoner, \jev{} leads all seven at this budget (Appendix~\ref{app:perreasoner}).

\par\vspace{2mm}
\noindent\textbf{Overall versus cut-only accuracy.}
Natural answers are shared across policies and dilute the differences between answerers.
On questions actually cut at 1.5k tokens, the mean \jev{} advantage is 8.6pp over forced answering, 6.2pp over \luna{} decision, and 6.9pp over \luna{} generative. At 32k tokens, the forced policy is 0.05pp higher overall. At 64k tokens, all four are within 0.1pp. Their convergence mainly reflects fewer cuts, not equivalent ability to use unfinished reasoning. Some runs never return a natural answer, so complete convergence is not required even at the largest budget.

\begin{table}[t]
\centering\small
\setlength{\tabcolsep}{3pt}
\resizebox{\columnwidth}{!}{%
\begin{tabular}{l c c cccc}
\toprule
 & & & \multicolumn{4}{c}{$B=1.5$k} \\
\cmidrule(lr){4-7}
Benchmark & Full & \jev{} $B{=}0$ & \jev{} & L-d & L-g & Forced \\
\midrule
GPQA-D     & 88.3 & 78.3 & 82.0 & 75.5 & 75.1 & 72.9 \\
SuperGPQA  & 66.3 & 56.2 & 64.7 & 60.3 & 59.6 & 59.7 \\
MedQA      & 95.1 & 88.6 & 95.0 & 94.9 & 94.9 & 94.1 \\
MedXpertQA & 48.9 & 32.2 & 49.9 & 49.3 & 48.9 & 47.5 \\
MMLU-Pro   & 86.1 & 84.5 & 87.4 & 84.5 & 84.0 & 84.3 \\
\bottomrule
\end{tabular}}
\caption{Accuracy (\%), mean over seven reasoners per benchmark. L-d/L-g: \luna{} decision/generative. $B=0$ is question-only. $B=1.5$k includes the shared natural-answer branch. SuperGPQA is weighted by the full benchmark's difficulty distribution.}
\label{tab:bench}
\end{table}

\par\vspace{2mm}
\noindent\textbf{Task and progress matter.}
At 1.5k, \jev{} exceeds forced answering by 9.2pp on GPQA-Diamond and 5.0pp on SuperGPQA, but by 0.9pp on MedQA and 2.4pp on MedXpertQA (Table~\ref{tab:bench}). On both medical benchmarks it is within one point of the two \luna{} interfaces. The percentage curves (Figure~\ref{fig:answerers}, bottom) remove the natural-answer branch and show how reader accuracy changes with relative saved-trace progress. RQ3 examines these changes against each reader's question-only baseline.
The two \luna{} interfaces differ by less than one point in macro-accuracy. Thus, the decision interface alone does not reproduce \jev{}'s advantage in this control, and the comparison does not isolate the roles of model, training or backend differences.

\par\vspace{2mm}
\noindent\textbf{Controlling the reasoning window.}
The main comparison gives forced answering the full prefix, whereas readers receive its final 1k tokens. We therefore conduct a post-hoc input-window control at $B=1.5$k on 6{,}500 matched cut outcomes covering 1{,}589 unique questions.
Both answerers receive either the full prefix or the identical saved 1k window, retaining their original answer-elicitation protocols.
The mean \jev{}--forced gap is 8.43pp with full prefixes (exploratory 95\% CI $[7.05, 9.85]$) and 7.96pp with matched 1k windows ($[6.52, 9.44]$ in Table~\ref{tab:windowcontrol}).
Intervals use 20{,}000 paired question-cluster bootstrap resamples, sharing question multiplicities across reasoners and stratifying SuperGPQA by difficulty. Thus, the observed gap persists under matched input windows. Model capabilities and answering interfaces remain different, so this control does not isolate an architectural effect.

\begin{table}[t]
\centering\small
\setlength{\tabcolsep}{5pt}
\begin{tabular}{l l r}
\toprule
Answerer & Reasoning window & Acc. (\%) \\
\midrule
Reasoner (forced) & Full prefix & 56.41 \\
Reasoner (forced) & Final 1k    & 57.01 \\
\jev{}           & Full prefix & 64.84 \\
\jev{}           & Final 1k    & 64.97 \\
\bottomrule
\end{tabular}
\caption{$B=1.5$k. Accuracy is macro-averaged over 35 pairs on the same 6{,}500 exact-cut outcomes. Naturally completed runs are excluded. Final 1k denotes the saved window of approximately 1{,}000 reasoning tokens. Acquired unparseable answers count as incorrect.}
\vspace{-4mm}
\label{tab:windowcontrol}
\end{table}

\par\vspace{2mm}
\noindent\textbf{Robustness and label sensitivity.}
Restricting the main fixed-budget comparison to the five strict-protocol reasoners gives a 2.9pp overall gap, smaller than the all-reasoner 4.1pp gap (Appendix~\ref{app:protocol}). For \jev{} versus forced answering at $B=1.5$k, Holm-adjusted percentile-bootstrap tests with 100{,}000 resamples show significantly higher accuracy for \jev{} in 17 of 35 pairs and significantly lower accuracy in none. Paired tests with the same 35-comparison Holm correction identify 14 higher-accuracy pairs and none lower (Appendix~\ref{app:multiple}). The ordering also holds on the 15{,}309 runs with a valid natural answer (Appendix~\ref{app:answered}).

\label{sec:noise}
Removing the 140 items flagged for model--label disagreement or manual review leaves a 4.4pp gap, with higher accuracy for \jev{} in 21 pairs and lower accuracy in none under the uncorrected 95\% paired-bootstrap interval criterion (Appendix~\ref{app:noise}). This is a sensitivity analysis, not a validated correction of benchmark labels.

\subsection{RQ2: Static budgets and adaptive stops}
\label{sec:cost}
\label{sec:stop}

\paragraph{Fixed-budget trade-offs.}
At $B=1.5$k, \jev{} costs an estimated \$1.29 per 1{,}000 questions, compared with \$1.46 for forced answering and \$4.44 for the full run, averaging these costs over pairs. Its cost ratio is 34.5\% of full, versus 38.3\% for forced answering, when ratios are averaged by pair. The ratio of the mean costs is instead 29.2\%. The corresponding accuracy changes against full are -1.13pp and -5.25pp. Estimated latency for \jev{} is 31.5\% of the full run, averaged over pairs (Appendix~\ref{app:latency}). These replay estimates do not establish online cancellation savings.

The 1.5k display point is chosen after inspecting the curve, not established as a universal optimum. Moving to 2k reduces the mean loss to 0.68pp while raising cost to 39.5\% of full. At 1k, the loss is 1.72pp and cost is 28.2\%. These averages indicate task differences: at 1.5k, \jev{} is 6.3pp below full on GPQA-Diamond and 1.6pp below on SuperGPQA, but 1.0pp above on MedXpertQA and 1.3pp above on MMLU-Pro. On answered runs alone, its mean loss is 1.4pp (Appendix~\ref{app:answered}).

On four matched subsets, standalone low-effort \luna{} costs less than all seven reasoner--\jev{} policies at $B=1.5$k and is at least as accurate in 18 of 28 comparisons (Appendix~\ref{app:nativeeffort}).
Savings against full runs by themselves therefore do not establish an advantage over native effort control.

\par\vspace{2mm}
\noindent\textbf{Tuned caps and adaptive increments.}
Out-of-fold selection chooses a separate static budget per pair and saves 63.3\% at a mean change of $-0.13$pp (Table~\ref{tab:stop}). This is not the uniform 1.5k policy.
The adaptive comparisons use 500-token checkpoints up to 16k, a 1k reading window and fold-selected budgets. They do not use the doubling schedule explored elsewhere. Every answering and signal-check call is charged. Question-only selection and cascades are additional controls in Appendix~\ref{app:necessity}, not substitutes for the matched cap comparison.

\begin{table}[t]
\centering\small
\setlength{\tabcolsep}{3.2pt}
\resizebox{\columnwidth}{!}{%
\begin{tabular}{l cc cc}
\toprule
 & \multicolumn{2}{c}{vs.\ full run} & \multicolumn{2}{c}{vs.\ tuned cap} \\
\cmidrule(lr){2-3}\cmidrule(lr){4-5}
Policy & Saving & $\Delta$Acc & $\Delta$Saving (+/$-$) & $\Delta$Acc ($-$) \\
\midrule
Tuned cap            & 63.3 & $-0.13$ & --              & --              \\
conf+stable          & 68.8 & $-0.42$ & $+5.5$ (14/9)   & $-0.28$ (3) \\
settled+useful       & 68.7 & $-0.61$ & $+5.4$ (19/1)   & $-0.47$ (4) \\
either               & 70.8 & $-0.62$ & $+7.5$ (20/0)   & $-0.49$ (5) \\
\textbf{auto}        & 71.3 & $-0.58$ & $+8.0$ (22/0)   & $-0.45$ (5) \\
\bottomrule
\end{tabular}}
\caption{Stop rules with \jev{} answering, out-of-fold selection, mean over 35 pairs. Saving in \% of full run's cost, and differences in points.
In brackets: pairs significantly better/worse (saving) or significantly worse (accuracy). Checks every 500 tokens, where every \jev{} call is charged.
\emph{auto} selects the rule in the training folds.}
\vspace{-4mm}
\label{tab:stop}
\end{table}

Relative to the tuned cap, conf+stable saves another 5.5 cost-percentage points at an additional 0.28pp accuracy loss, but is significantly more expensive in nine pairs. Settled+useful saves 5.4 points at an additional 0.47pp loss, and is significantly more expensive in one pair. Selecting the family within the training folds (\emph{auto}) saves 71.3\% overall: 8.0 points more than the tuned cap, at an additional 0.45pp accuracy loss. It is significantly cheaper in 22 pairs, with significant accuracy losses in five. These are jointly selected policies, so the difference includes both earlier stopping and changes in selected budgets.

\par\vspace{2mm}
\noindent\textbf{What actually stops the run?}
For \emph{auto}, 27.3\% of outcomes end at a rule-triggered stop, 37.9\% at the cap and 34.7\% by natural completion, averaged over pairs and CV repeats. It makes 1.52 \jev{} calls per question, including final answering (Appendix~\ref{app:deploy}). Thus, a substantial share of the efficiency gain is available without an adaptive trigger. The incremental saving is concentrated on GPQA-Diamond ($+21.0$ points) and SuperGPQA ($+12.6$), versus about three points on the medical benchmarks and
zero on MMLU-Pro, where the selected policies coincide.

\par\vspace{2mm}
\noindent\textbf{Accuracy losses on held-out folds.}
Policy selection limits accuracy loss to 1pp relative to the full run on training folds, but this constraint may not hold on held-out folds. On GPQA-Diamond, \emph{auto} loses 2.59pp versus 0.83pp for the tuned cap, while saving 59.6\% versus 38.6\% of full-run cost. Its additional saving therefore comes with a further 1.76pp accuracy loss.

Disabling early stopping while retaining the selected \emph{auto} budgets reduces the loss to 1.35pp. Thus, the early-stop rule adds 1.24pp of loss at the same budgets. This GPQA-only post-hoc diagnostic does not reselect policies (Appendix~\ref{app:gpqaaudit}).
Every \emph{auto} training fold had a feasible candidate, ruling out the no-feasible-candidate fallback as the source of the held-out loss.

\subsection{RQ3: Contribution of the reasoning trace}
\label{sec:prior}

\paragraph{A paired diagnostic, not a knowledge test.}
We compare each reader's percentage-cut answer with its question-only answer. This diagnostic retains 15{,}030 model--question outcomes covering 2{,}182 unique benchmark questions, requiring parseable answers from all three readers at 0, 10, 30, 50, 80 and 100\%. It excludes 347 outcomes that fail this condition, unlike the main evaluation where refusals and unparseable answers count as wrong. The following pooled percentages are unweighted, including SuperGPQA, and descriptive. Several reasoners contribute traces for the same question. They should not be read as the macro-averages above
or as 15{,}030 independent questions.

\begin{table}[t]
\centering\small
\setlength{\tabcolsep}{4pt}
\begin{tabular}{l r rrr}
\toprule
Reader & Cut (\%) & Fix (\%) & Hurt (\%) & Net (pp) \\
\midrule
\jev{}       & 10  & 6.2  & 3.7 & $+2.5$ \\
             & 50  & 12.0 & 4.9 & $+7.0$ \\
             & 100 & 14.4 & 5.2 & $+9.2$ \\
\luna{} dec. & 10  & 8.9  & 5.3 & $+3.5$ \\
             & 50  & 17.9 & 5.5 & $+12.4$ \\
             & 100 & 24.7 & 4.9 & $+19.8$ \\
\luna{} gen. & 10  & 10.2 & 7.4 & $+2.8$ \\
             & 50  & 17.5 & 6.3 & $+11.3$ \\
             & 100 & 24.2 & 5.1 & $+19.1$ \\
\bottomrule
\end{tabular}
\caption{Trace contribution on the common parseable-answer group ($n=15{,}030$ outcomes). Fix: question-only wrong, trace-conditioned right. Hurt: the reverse. Both divide by all included outcomes. Net gain equals fix minus hurt before rounding. A 100\% cut supplies only the final 1k reasoning window. Conditional rates are presented in Appendix~\ref{app:priorcontrol}.}
\vspace{-4mm}
\label{tab:tracegain}
\end{table}

\par\vspace{2mm}
\noindent\textbf{Net gains conceal both fixes and harms.}
\jev{} starts at 66.7\% question-only accuracy on this cohort, versus 56.0\% for \luna{} decision and 56.5\% for \luna{} generative. At a 10\% cut, its net gain of 2.5pp comprises 6.2\% fixes and 3.7\% harms. At 100\%, 14.4\% fixes and 5.2\% harms yield a 9.2pp gain (Table~\ref{tab:tracegain}).
\luna{} gains more from its lower starting point, so the readers' accuracies approach one another. Neither the higher starting accuracy nor the smaller net gain identifies how a model internally uses knowledge or reasoning.

\par\vspace{2mm}
\noindent\textbf{Different tasks benefit at different stages.}
\label{sec:where}
\jev{}'s net gain at 10\% is $-0.4$pp on GPQA-Diamond, $+0.2$pp on SuperGPQA and $+0.3$pp on MMLU-Pro, but $+8.6$pp on MedXpertQA (Appendix~\ref{app:priorcontrol}). GPQA-Diamond increases to $+10.9$pp at 100\%, indicating that later trace windows can be substantially more useful than early ones. This complements the fixed-budget result: short cuts do not preserve GPQA accuracy even when \jev{} outperforms other answerers at those cuts. Relative progress is available only retrospectively, so these observations do not specify a deployable stopping rule.

\par\vspace{2mm}
\noindent\textbf{Controlling the question-only starting point.}
Among 3{,}455 outcomes where all three readers start wrong, their accuracies at 100\% are 34.2\%, 34.0\% and 33.7\%, respectively. These similar descriptive rates do not establish equivalence. Among 6{,}597 outcomes where all start right, 97.8\%, 96.5\% and 94.6\% remain correct at 10\%. The observed advantage therefore includes retaining correct initial answers as well as recovering from wrong ones. Comparing conditional fix rates alone would confound different starting error sets (Appendix~\ref{app:priorcontrol}).

\par\vspace{2mm}
\noindent\textbf{Completed traces as a boundary case.}
\label{sec:complete}
On the 15{,}309 runs with a valid natural answer, \jev{} reading the final approximately 1k reasoning window selects the same option as the reasoner on 99.1\% on average across pairs. Its mean accuracy is 0.4pp higher, with 7 pairs having an uncorrected 95\% paired-bootstrap interval entirely above zero and none below (Appendix~\ref{app:complete}). High agreement does not establish copying or faithful verification.

\section{Discussion}
\label{sec:discussion}
\vspace{-1mm}

Our evaluation yields three main findings. \textbf{First, Jev's advantage persists under matched input windows.} The fixed-cut comparisons show that answerer choice affects accuracy even without a learned stopping policy. On the exact-cut control cohort at $B=1.5$k, with 6{,}500 matched cuts, \jev{} retains a 8.0pp macro-average advantage over forced answering when both receive the identical final 1k window (Section~\ref{sec:same}). The shorter reading window therefore does not alone explain the observed gap. These results support evaluating who answers separately from when reasoning stops. However, model capabilities and answer-elicitation interfaces remain different, so the comparison does not isolate the System-1 architectural advantage.

\textbf{Second, adaptive savings come with task-dependent accuracy losses.} Tuned caps provide substantial savings without intermediate checks. Joint adaptive selection adds 8.0 cost-percentage points of saving at a further 0.45pp accuracy loss on average, with larger losses on GPQA-Diamond (Section~\ref{sec:cost}). Its benefit is therefore a task-dependent trade-off rather than a uniform improvement over static budgets. The GPQA results also show that a training-fold loss constraint is not a deployment guarantee. Budget, answerer and checking frequency should be evaluated against a strong static-budget baseline and the accuracy loss an application can tolerate.

\textbf{Third, trace gains depend on the reader, task and stage.}
On the paired cohort, \jev{} has higher question-only accuracy,
while \luna{} gains more from traces (Section~\ref{sec:prior}).
Within medicine, \jev{} starts high on MedQA but low on MedXpertQA, where early windows yield substantial gains.
MMLU-Pro shows smaller increments, whereas GPQA-Diamond benefits from later windows despite high starting accuracy.
These patterns motivate evaluating starting accuracy and trace benefit separately, rather than attributing benchmark differences to discipline alone.
They describe behavioral changes, not internal mechanisms,
and do not establish reasoning faithfulness. Percentage cuts are retrospective diagnostics, but not deployable stopping rules.

\section{Conclusion}
\label{sec:conclusion}
We evaluate System-1 decision models as decision-makers for answer selection from truncated reasoning across seven LLM reasoners and five benchmarks. At a 1,500-token budget, \jev{} outperforms forced answering by 4.1 percentage points on average at about one-third of full-run cost. Adaptive stopping offers further savings, with accuracy losses that vary by task. Question-only controls show that readers' standalone accuracy and their gains from reasoning traces both matter, while longer reasoning remains valuable on harder tasks. These findings highlight answerer choice alongside reasoning budget as a way to improve reasoning efficiency.

\section*{Limitations}
\vspace{-1mm}

This study presents several directions for future research. First, we study multiple-choice tasks, one \jev{} system and one general-purpose control model through two interfaces. The findings do not establish advantages for all decision models, open-ended tasks or readers allowed additional reasoning. Same cut does not mean same input. Readers see an approximately 1k-token window, while forced answering generally receives the whole prefix. Forced-answer protocols also differ across reasoners, where an alternative protocol may perform better. The strict-only 1.5k gap is 2.9pp rather than 4.1pp. \luna{}'s two interfaces need not have identical backend implementations.

The reported policies are evaluated by offline replay, not by interrupting the saved run online. Costs use recorded usage and historical list prices, and latency is modelled from generation speeds and answer-call latencies. Neither measures cancellation billing, asynchronous overhead or actual deployment savings. The approach requires visible reasoning text, which need not reveal all internal computation. Token-to-text mapping is sometimes approximate. Forced answers are matched to the saved cut at 1.5k, and other budgets retain the nearest earlier collected position when exact answers are unavailable (Appendix~\ref{app:exact}). Percentage cuts require the completed trace length and are diagnostic, not online policies.

The answerer comparison follows earlier analyses of these traces, and the 1.5k display point is post-hoc. Out-of-fold selection separates cap and threshold fitting from evaluation, but does not remove research-design selection effects or guarantee a one-point held-out loss. The four historical MMLU-Pro checks used an earlier protocol and are not independent confirmation (Appendix~\ref{app:heldout}). Per-pair intervals are mostly uncorrected. The shared questions and reasoners make comparisons dependent, and nonsignificance does not establish equivalence. RQ3 pools repeated model--question outcomes and excludes 347 outcomes lacking a parseable answer at any required point, so its descriptive rates may reflect selection effects. The GPQA diagnostic intervals condition on the recorded policy choices rather than refitting selection in each bootstrap.

\paragraph{Data and reproducibility.}
Traces and forced answers are single samples requested at temperature 1.0, not repeated estimates of generation variability. The 408 retained 64k reruns are fresh, selectively collected samples, but other runs retain their original limits. The 67 retained no-answer runs and 1 saved failed-generation outcome count as wrong for full reasoning, as do reader refusals and unresolved parsing failures in the main evaluation.
Public-benchmark exposure and label errors remain possible, but model--label disagreement is not proof that a gold label is wrong. SuperGPQA main results are weighted estimates from a difficulty-balanced sample, unlike the unweighted RQ3 diagnostic.
Hosted models and prices can change: \jev{} requests used the \texttt{jev-latest} alias, although the audited budget responses all reported \texttt{jev-1.13.0}. Saved outputs document the evaluated snapshot, not a guarantee of future reproducibility.

\section*{Statement of LLM Use}
\label{app:use_llm}

During the preparation of this paper, the authors used LLM-based tools, primarily Claude and ChatGPT, to assist in code development and debugging, improve grammar and clarity in draft text, and refine the layout of schematic figures. The authors reviewed the generated suggestions and take full responsibility for the final code, analyzes, figures, and manuscript.

\bibliography{reference}

\appendix

\twocolumn[{%
\section*{Appendix Contents}
\label{app:contents}
\vspace{-1mm}

\etocsetlocaltop.toc{part}
\etocsetnexttocdepth{subsection}
\etocstandardlines
\etocsettocstyle{}{}
\expandafter\renewcommand\csname l@section\endcsname{\csname @dottedtocline\endcsname{1}{0em}{1.5em}}
\expandafter\renewcommand\csname l@subsection\endcsname{\csname @dottedtocline\endcsname{2}{1.5em}{2.8em}}
\localtableofcontents
\medskip
}]

\section{Prompts and request settings}
\label{app:prompts}
{\sloppy
\paragraph{LLM Reasoners.} All seven reasoners are called through the Together AI API between late September and early October 2026 with thinking enabled. The collection scripts requested temperature 1.0 and a 32{,}000-token output limit (64{,}000 for reruns):

\begin{table}[!t]
\centering
{\small\setlength{\tabcolsep}{3pt}\resizebox{\columnwidth}{!}{\begin{tabular}{@{}ll r@{}}
\toprule
Reasoner & Model identifier & In / out (\$/1M) \\
\midrule
DeepSeek V4 Pro   & \texttt{deepseek-ai/DeepSeek-V4-Pro-0813}   & 1.32 / 3.96 \\
Qwen3.7 Plus      & \texttt{Qwen/Qwen3.7-Plus}                  & 0.32 / 1.28 \\
MiniMax M3        & \texttt{MiniMaxAI/MiniMax-M3}               & 0.30 / 1.20 \\
GPT-OSS 120B      & \texttt{openai/gpt-oss-120b} (high)         & 0.15 / 0.60 \\
GLM-5.3 Flash     & \texttt{zai-org/GLM-5.3-Flash}              & 0.15 / 0.50 \\
Qwen3.8 Flash     & \texttt{Qwen/Qwen3.8-Flash}                 & 0.15 / 0.47 \\
DeepSeek V4 Flash & \texttt{deepseek-ai/DeepSeek-V4-Flash-0731} & 0.14 / 0.28 \\
\bottomrule
\end{tabular}}}
\caption{Seven LLM reasoners used in this study.}
\label{tab:reasoners}
\end{table}

The full-run prompt is:
\begin{promptcard}{pcgrey}{Reasoner, full run}
Answer the following multiple-choice question. Think step by step, then end your response with a single line of the form \textquotesingle Answer: X\textquotesingle{} where X is one of \pslot{option letters}.
\pgap
Question:\\
\pslot{question}
\pgap
Options:\\
A. \pslot{option A}\\
B. \pslot{option B}\\
...
\end{promptcard}

\paragraph{Forced answer, strict.} A new chat request with thinking switched off (DeepSeek
\texttt{reasoning.enabled=false}; Qwen \texttt{enable\_thinking=false}; MiniMax
\texttt{thinking\_mode=disabled}), at most 48 output tokens; system and user messages:
\begin{promptcard}{pcforced}{Forced answer, strict}
\pcpart{System message}
You are being asked for an immediate final answer. Do not solve, explain or show any work. Reply with exactly two lines and nothing else:\\
Answer: <letter>\\
Confidence: <0-100>
\pcpart{User message}
\pslot{full-run prompt}
\pgap
Below is your own reasoning so far on this question. It was cut off before you finished.
\pgap
<reasoning>\\
\pslot{prefix}\\
</reasoning>
\pgap
Based on this reasoning, give your best answer now and how confident you are that it is correct. Reply with exactly two lines:\\
Answer: X\\
Confidence: N\\
where N is an integer from 0 to 100.
\end{promptcard}

\paragraph{Forced answer, think-close.} A raw completion request in the model's template: the
full-run prompt as the user turn, the prefix as the model's thinking, the thinking block closed, and the
answer started with the stub \texttt{Answer:}; 4 output tokens, the first option letter is taken.
Templates were read from the model repositories on 4 October 2026.
\begin{promptcard}{pcforced}{Forced answer, think-close}
\pcpart{GLM-5.3 Flash}
[gMASK]<sop>\\
<|system|>Reasoning Effort: Max\\
<|user|>\pslot{full-run prompt}\\
<|assistant|><think>\pslot{prefix}\allowbreak</think>\pstub{Answer:}
\pcpart{GPT-OSS 120B (harmony format)}
<|start|>system\allowbreak<|message|>You are ChatGPT, a large language model trained by OpenAI.\pnl\allowbreak Knowledge cutoff: 2024-06\pnl\pnl\allowbreak Reasoning: high\pnl\pnl\allowbreak \# Valid channels: analysis, commentary, final. Channel must be included for every message.\allowbreak<|end|>\\
<|start|>user\allowbreak<|message|>\pslot{full-run prompt}\allowbreak<|end|>\\
<|start|>assistant\allowbreak<|channel|>analysis\allowbreak<|message|>\pslot{prefix}\allowbreak<|end|>\\
<|start|>assistant\allowbreak<|channel|>final\allowbreak<|message|>\pstub{Answer:}
\pcnote{Each prompt is one string. Lines start at turn boundaries for reading; \textbackslash n marks a newline inside the prompt.}
\end{promptcard}

\paragraph{\jev{}.} Requested as \texttt{jev-latest}; retained budget responses identify \texttt{jev-1.13.0}. Endpoint: \texttt{POST /v1/systemone}. State: question, lettered options, and
\texttt{partial\_reasoning} (the last 1{,}000 tokens, prefixed with ``\ldots'' when cut; ``(no reasoning
yet)'' when empty). Three typed questions in one call:
\begin{promptcard}{pcjev}{\jev{}, typed questions}
\pcpart{answer (choice; criteria = option texts cut to 300 characters)}
The state contains a multiple-choice question and a reasoner\textquotesingle s partial chain of thought (it may be cut off mid-sentence, or empty). Which option is the correct answer?
\pcpart{settled (yes/no probability)}
Has the partial reasoning already clearly committed to a final answer, so that continuing to think is unlikely to change it?
\pcpart{useful (yes/no probability)}
If the reasoner keeps thinking, is it still likely to change or correct its current answer (for example: unresolved steps, open doubts, or a pending check that could flip it)?
\end{promptcard}
\noindent With $B=0$ only the answer question is asked, with ``The state contains a multiple-choice question''.

\paragraph{\luna{}.} \texttt{gpt-6-luna}. Decisions API (\texttt{POST /v1/decisions}, public beta): the same
question, options and window as text, the \jev{} answer instruction (``The input contains \ldots''), one choice
per option letter with the option text (300 characters) as description; no fallback option.
Generative (\texttt{/v1/responses}, reasoning effort \texttt{none}, at most 48 output tokens, 512 on the single
retry): the \jev{} answer instruction followed by the question, options and window, with the strict
forced-answer system message and two-line output format.\par}

\paragraph{Reader prices and request latencies.}
\jev{} costs \$0.042 per million input tokens, with median recorded latency 0.13\,s over 92{,}763 unique budget calls.
Luna Decisions costs \$0.10 per million input tokens, with median initial-call latency 0.14\,s; Luna generation has median latency 0.96\,s.
Luna medians exclude re-asks and are acquisition statistics, not matched-window comparisons with \jev{}.
All prices are recorded list prices, rather than independently verified per-request bills.
The Luna interface comparison does not assume identical backend implementations.

\section{Data and completeness}
\label{app:data}
Table~\ref{tab:pairs} summarizes sample counts, reruns, answer availability, and reasoning lengths for each reasoner--benchmark pair.

\begin{table}[h]
\centering\small
\setlength{\tabcolsep}{2.5pt}
\resizebox{\columnwidth}{!}{%
\begin{tabular}{ll rrr rr}
\toprule
Reasoner & Benchmark & $n$ & Rerun & No final & Median & Finish \\
 & & & 64k & answer & tokens & $\le$1.5k (\%) \\
\midrule
DeepSeek V4 Pro & GPQA-D & 198 & 9 & 1 & 852 & 59 \\
 & SuperGPQA & 500 & 5 & 1 & 501 & 75 \\
 & MedQA & 500 & 0 & 0 & 195 & 94 \\
 & MedXpertQA & 500 & 1 & 0 & 725 & 67 \\
 & MMLU-Pro & 500 & 3 & 0 & 172 & 86 \\
Qwen3.7 Plus & GPQA-D & 198 & 7 & 0 & 3{,}401 & 19 \\
 & SuperGPQA & 500 & 16 & 1 & 1{,}547 & 49 \\
 & MedQA & 500 & 0 & 0 & 1{,}095 & 71 \\
 & MedXpertQA & 500 & 9 & 0 & 1{,}984 & 28 \\
 & MMLU-Pro & 500 & 1 & 0 & 997 & 69 \\
MiniMax M3 & GPQA-D & 198 & 17 & 4 & 2{,}948 & 33 \\
 & SuperGPQA & 500 & 24 & 4 & 2{,}084 & 45 \\
 & MedQA & 500 & 5 & 1 & 603 & 82 \\
 & MedXpertQA & 500 & 9 & 1 & 2{,}838 & 31 \\
 & MMLU-Pro & 500 & 7 & 1 & 654 & 68 \\
GPT-OSS 120B & GPQA-D & 194 & 13 & 0 & 4{,}005 & 26 \\
 & SuperGPQA & 496 & 13 & 2 & 2{,}738 & 32 \\
 & MedQA & 500 & 0 & 0 & 683 & 79 \\
 & MedXpertQA & 500 & 0 & 0 & 2{,}114 & 33 \\
 & MMLU-Pro & 499 & 2 & 0 & 1{,}041 & 62 \\
GLM-5.3 Flash & GPQA-D & 198 & 22 & 8 & 4{,}538 & 33 \\
 & SuperGPQA & 500 & 31 & 7 & 2{,}045 & 43 \\
 & MedQA & 500 & 2 & 0 & 636 & 70 \\
 & MedXpertQA & 500 & 10 & 0 & 6{,}719 & 15 \\
 & MMLU-Pro & 500 & 6 & 0 & 469 & 70 \\
Qwen3.8 Flash & GPQA-D & 198 & 34 & 9 & 6{,}089 & 27 \\
 & SuperGPQA & 500 & 42 & 7 & 2{,}981 & 39 \\
 & MedQA & 500 & 4 & 0 & 535 & 72 \\
 & MedXpertQA & 500 & 50 & 4 & 7{,}833 & 15 \\
 & MMLU-Pro & 500 & 14 & 0 & 435 & 69 \\
DeepSeek V4 Flash & GPQA-D & 198 & 11 & 3 & 1{,}053 & 56 \\
 & SuperGPQA & 500 & 23 & 5 & 471 & 70 \\
 & MedQA & 500 & 4 & 1 & 200 & 93 \\
 & MedXpertQA & 500 & 6 & 3 & 718 & 64 \\
 & MMLU-Pro & 500 & 8 & 5 & 171 & 85 \\
\bottomrule
\end{tabular}}
\caption{Per pair: retained runs, 64k reruns, outcomes without a valid natural answer (including generation failure, scored incorrect for the full run), median reasoning tokens, and share with a valid natural answer within 1{,}500 tokens.}
\label{tab:pairs}
\end{table}
The combined 64k rerun cohort contains 416 of 15{,}385 original runs:
413 have a \texttt{length} finish reason, and three lack valid answers despite other finish reasons.
Each rerun generates a fresh sample rather than continuing the original trace. Failed requests were sometimes retried.
Eight failed rerun requests are excluded, leaving 408 rerun traces: 341 with valid final answers and 67 without (59 with no option letter and eight with a letter only in truncated output).
Ten requests that dropped the connection were repeated, with up to three attempts.
The retained acquisition cohort contains 15{,}377 runs: 15{,}309 have valid natural answers, 67 are the no-valid-answer rerun outcomes described above, and one additional saved full-generation stream ended with \texttt{finish\_reason=error}. The failed stream is retained and scored incorrect, not classified as a natural completion. At $B=1.5$k, 15{,}376 runs enter the four-way comparison because one exact forced answer remains unavailable. The two filtered forced-window requests for that outcome are treated as missing rather than scored as incorrect.
Within each comparison, all answerers and the full-run baseline use the same retained original or rerun trace. Selective rerun replacement is not an unconditional sample under a uniform 64k limit.
For client-capped Qwen runs, the 32k and 64k limits are approximated by 115{,}200 and 230{,}400 reasoning characters, respectively.
GPQA-Diamond, MedQA, MedXpertQA-Text and MMLU-Pro questions are used as released; the SuperGPQA subset contains 167 easy, 166 middle, and 167 hard questions and is reweighted to the full dataset's difficulty shares.
Three requests of Qwen3.8 Flash on one SuperGPQA question are rejected by the provider's input filter and count as wrong; no question is rephrased.
\paragraph{Luna acquisition.}
Deduplicated initial acquisition logs contain 191{,}637 calls per interface, covering question-only,
budget and percentage probes before final evaluation-cohort filtering.
There are 1{,}613 nonblocked generative parse failures (0.84\%) and two blocked calls;
the decision interface returns 1{,}107 refusals (0.58\%).
These are acquisition statistics, not final-cohort refusal rates.
Refusals, blocked calls, and parse failures remaining after the configured re-ask count as incorrect.

\paragraph{Retry sensitivity.}
The retry increased \luna{} generative accuracy at the 100\% cut from 74.0\% to 77.0\%. At all other cuts, accuracy changed by at most 0.4 percentage points.


\section{Replay implementation and cost accounting}
\label{app:methoddetails}

\subsection{Cut positions and reader windows}
Positions use saved tokenizer offsets where available. Otherwise the nominal token position is mapped
proportionally to a character offset and snapped back to a newline within 200 characters.
Consequently the 1{,}000-token reader window is nominal, not an exact token count for every trace.
At $B=1.5$k, 615 additional full-prefix forced answers complete exact-position coverage for 6{,}500 matched cut outcomes. One of the 6{,}501 target outcomes remains missing because both forced-window requests were filtered. It is excluded from matched comparisons. Other budgets retain the earlier-position fallback when an exact forced answer is unavailable. Appendix~\ref{app:exact} reports the updated coverage and sensitivity analysis.

\paragraph{Relative-progress diagnostics.}
Cuts use $c_f=\operatorname{round}(fL)$ for $f\in\{0.1,0.2,\ldots,1.0\}$.
Every point invokes an answerer, including at 100\%, where the input contains completed reasoning but excludes separately generated final-answer content.
Because $L$ is known only after generation, these are retrospective diagnostics.

\subsection{Cost accounting details}
\label{app:cost-accounting}
\paragraph{Reasoner costs.}
Natural completions are charged for the original prompt and all generated output, including the final answer.
Truncated runs are charged for the prompt and reasoning through the cut.
Partial output usage is estimated from the saved trace, rather than a cancelled request.

\paragraph{Answerer costs.}
Reader inputs include instructions, the question, options, and the reasoning window.
\jev{} and Luna Decisions incur input charges only.
Luna generation incurs input and output charges, including required re-asks.
Forced answering incurs a separate call charged at the reasoner's input and output prices, with cached input charged at the cached rate.

\paragraph{Usage, acquisition, and latency.}
Recorded request usage supplies token counts where available.
Missing Luna Decisions input usage is estimated from serialized request length.
Repeated windows are charged at every deployed check, even when acquisition reused a response.
Acquiring full traces, probes, forced answers and unrelated failed retries incurs separate experimental expense, excluded from single-policy cost. Required answer-format re-asks belong to the selected answering chain and are included. Unknown usage is estimated and flagged, not treated as free. Latency estimation and its limitations are detailed in Appendix~\ref{app:latency}.

\subsection{Trace-contribution denominators}
A fix changes an incorrect question-only answer to a correct trace-conditioned answer, where a hurt is the reverse.
Net trace gain equals the fix rate minus the hurt rate when both use a common denominator.
Conditional fix and hurt rates instead divide by question-only errors and correct answers, respectively, and are not directly comparable.
Pooled counts are model--question outcomes because a question may have traces from several reasoners.
These outcomes are not independent questions.
Trace gain measures behavioral contribution, not an internal knowledge source or reasoning faithfulness.
Failure to detect an accuracy loss is not evidence of equivalence.

\section{Criteria fixed before four held-out cells}
\label{app:heldout}
Table~\ref{tab:heldout} reports the six pre-specified criteria and their outcomes under the earlier frozen protocol.
The comparison of answerers was found on 31 reasoner--benchmark cells.
Before the MMLU-Pro cells of DeepSeek V4 Pro, DeepSeek V4 Flash, MiniMax M3 and Qwen3.8 Flash had been seen, we fixed the analysis script (identified by its MD5 hash) and the six criteria below, and committed to report them whichever way they came out.
Reader: last 12{,}000 characters of the prefix, forced answers: each reasoner's main protocol, full runs with the original 32{,}000-token limit, paired bootstrap over questions, 2{,}000 resamples, 95\% intervals. an interval excludes zero only if its bound is beyond $10^{-9}$.
The formal check was run later: before it, the four cells had already entered aggregate figures and tables of later analyses.
The frozen script was run unchanged. Only its output folder, and for the two DeepSeek models the folder it reads
the MMLU-Pro data from, were set at run time. Its numbers are identical to those of the cell files produced on the day the data arrived.
The later changes (64k reruns, 1{,}000-token window, budget-level cost model) are not part of this check.

\begin{table*}[t]
\centering\small
\setlength{\tabcolsep}{4pt}
\begin{tabular}{p{5.4cm}cccc c}
\toprule
Criterion (all at $B=1{,}500$ unless stated) & DS V4 Pro & DS V4 Flash & MiniMax M3 & Qwen3.8 Flash & Result \\
\midrule
1. Reader $-$ forced on cut questions with an exact forced answer: positive in all four, CI excludes 0 in $\ge 2$
 & $+9.8^{*}$ & $+6.8$ & $+13.3^{*}$ & $+3.5$ & pass \\
 & \scriptsize $n=61$ & \scriptsize $n=59$ & \scriptsize $n=135$ & \scriptsize $n=113$ & \\
2. Fixed budget + reader not significantly below the full run in any cell
 & $+2.0^{*}$ & $+2.0^{*}$ & $+2.4^{*}$ & $+1.4$ & pass \\
3. Forced answers lose more than the reader (accuracy change vs full run, forced / reader)
 & $+0.8$ / $+2.0$ & $0.0$ / $+2.0$ & $-2.2$ / $+2.4$ & $-0.2$ / $+1.4$ & pass \\
4. Tuned cap (out of fold, $t=1$): reader saves more dollars than forced, CI excludes 0 in $\ge 3$; cap + reader not
 significantly below the full run
 & $+10.5^{*}$ / $+1.2$ & $+7.7^{*}$ / $+1.4$ & $+37.1^{*}$ / $+2.6^{*}$ & $+7.2^{*}$ / $+0.6$ & pass \\
5. Completed reasoning, limit-hit traces removed: reader within 1 point of the reasoner
 & $+0.4$ & $+0.2$ & $+1.2^{*}$ & $+0.2$ & \textbf{fail} \\
6. Stop rule + reader $-$ cap + reader, dollars: $|\text{mean}| < 3$
 & $0.0$ & $0.0$ & $0.0$ & $0.0$ & pass$^{\dagger}$ \\
\bottomrule
\end{tabular}
\caption{Pre-specified criteria on the four held-out MMLU-Pro cells. Differences in points;
$^{*}$ 95\% paired-bootstrap interval excludes 0. Row 4 gives the dollar-saving difference / the accuracy change of
cap + reader against the full run.
Criterion 5 fails because the reader on the completed reasoning of MiniMax M3 is 1.2 points \emph{more} accurate
than the reasoner's own answer [0.4, 2.4]; the stated tolerance was symmetric.
$^{\dagger}$Trivially: on all four cells the out-of-fold selection chose a cap of 500 tokens in every fold; the stop rule checks every 500 tokens, so it has no chance to stop before the cap and the two policies coincide.}
\label{tab:heldout}
\end{table*}

\section{Per-reasoner results and costs}
\label{app:perreasoner}
\begin{figure*}[t]
\centering
\includegraphics[width=\textwidth]{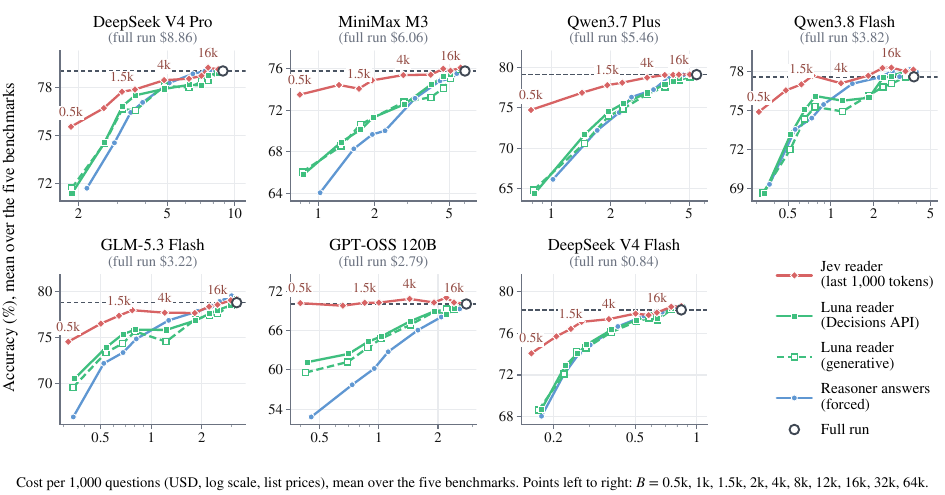}
\caption{Accuracy against cost per reasoner, mean over the five benchmarks, in dollars per 1{,}000
questions at list prices (log scale). Each line traces the budgets $B=0.5$k to 64k; labels mark \jev{}'s
points. Open circle: full run.}
\label{fig:cost}
\end{figure*}

Table~\ref{tab:perreasoner} reports accuracy and relative cost at $B=1.5$k for each reasoner, aggregated over the five benchmarks. In addition, Figure~\ref{fig:cost} shows accuracy--cost curves for each reasoner, averaged over the five benchmarks.

\begin{table}[h]
\centering\small
\setlength{\tabcolsep}{2.6pt}
\begin{tabular}{l c cccc cc}
\toprule
 & & \multicolumn{4}{c}{Accuracy (\%)} & \multicolumn{2}{c}{Cost (\%)} \\
\cmidrule(lr){3-6}\cmidrule(lr){7-8}
Reasoner & Full & \jev{} & L-d & L-g & F & \jev{} & F \\
\midrule
DeepSeek V4 Pro     & 79.0 & 77.7 & 76.8 & 76.7 & 76.5 & 35.3 & 38.7 \\
Qwen3.7 Plus        & 79.1 & 77.8 & 74.5 & 74.0 & 74.3 & 35.2 & 40.6 \\
MiniMax M3          & 75.8 & 74.1 & 70.1 & 70.6 & 69.3 & 27.2 & 32.2 \\
GPT-OSS 120B        & 70.0 & 70.2 & 64.3 & 63.4 & 59.9 & 30.3 & 34.1 \\
GLM-5.3 Flash       & 78.8 & 77.3 & 75.4 & 74.3 & 73.3 & 20.0 & 21.3 \\
Qwen3.8 Flash       & 77.6 & 77.0 & 75.1 & 74.4 & 74.3 & 16.2 & 19.4 \\
DeepSeek V4 Flash   & 78.2 & 76.4 & 74.0 & 74.2 & 74.3 & 28.9 & 32.0 \\
\bottomrule
\end{tabular}
\caption{Accuracy is averaged over five benchmarks at $B=1.5$k on the matched cohort. Cost is the ratio of mean policy dollars to mean full-run dollars across those benchmarks, unlike the main table's mean of within-pair ratios. Full-run cost per 1{,}000 questions, in row order: \$8.86, 5.46, 6.06, 2.79, 3.22, 3.83, 0.84.}
\vspace{-4mm}
\label{tab:perreasoner}
\end{table}


\section{Latency analysis}
\label{app:latency}

For a truncated run, latency is estimated as the retained reasoning length divided by the recorded full-run decoding speed, plus the answer-call latency. For natural finishes, we retain the full-run estimate without an additional answer call. Full-run latency is likewise estimated from reasoning length and decoding speed.

At $B=1.5$k, mean estimated latency is 16.5\,s for \jev{} and 72.6\,s for the full run across 35 reasoner--benchmark pairs. Averaging within-pair latency ratios gives 31.5\%; the ratio of mean times is instead 22.7\%. These summaries use unweighted question means within each pair and equal weights across pairs. 
The estimates do not separately account for time to first token, online cancellation, or asynchronous checking overhead. We therefore interpret them as decoding-based replay estimates rather than measured end-to-end response times.

\section{Native reasoning-effort baselines}
\label{app:nativeeffort}

\paragraph{Matched-question comparison.}
We compare standalone GPT-6 Luna at reasoning efforts
\textbf{none}, \textbf{low}, \textbf{medium}, and \textbf{high}
with the current fixed-budget reasoner--\jev{} policies at $B=1.5$k.
Standalone Luna receives the question and options without external
reasoning. The Jev policies retain natural answers within the budget
and otherwise use the final approximately 1k-token reasoning window.
All methods share the intersection of question IDs across the four
Luna settings and seven retained reasoner cohorts: 194 GPQA-Diamond,
297 SuperGPQA, 500 MedQA, and 500 MedXpertQA questions. The Luna
acquisition covers the earlier 300-question SuperGPQA sample and
does not cover MMLU-Pro. Refusals and unparseable answers count as
incorrect. SuperGPQA means use the full-dataset difficulty weights,
consistent with the main evaluation.

Table~\ref{tab:nativeeffort} reports accuracy and total estimated
cost at the recorded list prices. Jev policy costs include the
reasoner prompt, retained output, and reader call. Luna costs use
recorded input, cached input, and total output usage.

\begin{table*}[t]
\centering\small
\setlength{\tabcolsep}{4pt}
\begin{tabular}{l rr rr rr rr}
\toprule
 & \multicolumn{2}{c}{GPQA-D ($n=194$)}
 & \multicolumn{2}{c}{SuperGPQA ($n=297$)}
 & \multicolumn{2}{c}{MedQA ($n=500$)}
 & \multicolumn{2}{c}{MedXpertQA ($n=500$)} \\
\cmidrule(lr){2-3}\cmidrule(lr){4-5}\cmidrule(lr){6-7}\cmidrule(lr){8-9}
Method & Acc. & Cost & Acc. & Cost & Acc. & Cost & Acc. & Cost \\
\midrule
\luna{} none & 64.9 & 0.087 & 53.4 & 0.071 & 91.4 & 0.059 & 42.4 & 0.066 \\
\luna{} low & 85.1 & 0.305 & 65.0 & 0.218 & 95.0 & 0.113 & 53.4 & 0.197 \\
\luna{} medium & 88.1 & 0.629 & 69.0 & 0.355 & 95.4 & 0.147 & 53.2 & 0.329 \\
\luna{} high & 88.1 & 1.028 & 67.1 & 0.562 & 95.6 & 0.181 & 53.0 & 0.494 \\
\midrule
\multicolumn{9}{l}{Reasoner + \jev{}, $B=1.5$k} \\
DS V4 Pro & 84.5 & 4.272 & 68.2 & 3.218 & 95.6 & 1.876 & 53.0 & 3.946 \\
Qwen3.7 Plus & 84.5 & 2.100 & 67.6 & 1.871 & 95.2 & 1.885 & 55.4 & 2.094 \\
MiniMax M3 & 83.0 & 1.864 & 64.4 & 1.671 & 93.2 & 1.420 & 43.8 & 1.924 \\
GPT-OSS 120B & 75.3 & 0.974 & 53.1 & 0.916 & 94.4 & 0.646 & 42.2 & 0.957 \\
GLM-5.3 Flash & 85.1 & 0.738 & 68.5 & 0.657 & 95.0 & 0.534 & 51.6 & 0.821 \\
Qwen3.8 Flash & 83.0 & 0.732 & 67.5 & 0.648 & 96.0 & 0.463 & 52.0 & 0.793 \\
DS V4 Flash & 84.0 & 0.337 & 65.5 & 0.242 & 95.6 & 0.136 & 51.2 & 0.316 \\
\bottomrule
\end{tabular}
\caption{Standalone Luna effort baselines and fixed-budget Jev
policies on shared questions. Accuracy in \%; estimated cost in
USD per 1{,}000 questions. All rows use the same questions within
each benchmark, with difficulty weighting on SuperGPQA.
Point estimates are descriptive, without significance claims.}
\label{tab:nativeeffort}
\end{table*}

\paragraph{Effort and output usage.}
Table~\ref{tab:nativeefforttokens} reports Luna's total output
tokens, including hidden reasoning. Low effort substantially
reduces output usage relative to high effort on these subsets,
but the accuracy differences vary by benchmark. For example,
on MedXpertQA, low uses 331 output tokens per question at 53.4\%
accuracy, versus 924 tokens and 53.0\% for high. Similar point
estimates do not establish equivalence.

\begin{table}[t]
\centering\small
\setlength{\tabcolsep}{4pt}
\begin{tabular}{lrrrr}
\toprule
Effort & GPQA-D & SuperGPQA & MedQA & MedX. \\
\midrule
none & 123 & 89 & 66 & 69 \\
low & 559 & 384 & 176 & 331 \\
medium & 1207 & 658 & 243 & 593 \\
high & 2006 & 1071 & 311 & 924 \\
\bottomrule
\end{tabular}
\caption{Mean total output tokens for standalone Luna on the
matched subsets of Table~\ref{tab:nativeeffort}. Reasoning tokens
are included in output usage and are not added again.}
\vspace{-4mm}
\label{tab:nativeefforttokens}
\end{table}

\paragraph{Interpretation and scope.}
At $B=1.5$k, low-effort \luna{} has lower estimated cost in all
28 reasoner--benchmark comparisons and at least as high accuracy
in 18. Thus, the savings against full reasoner runs do not
establish superiority over native effort control or alternative
standalone models. This comparison changes the model as well
as the answering procedure; it does not isolate effort within
each of the seven reasoners or establish the best effort setting.
The comparisons share questions and are not independent replications.
These counts describe the 1.5k display budget, not tuned caps,
adaptive policies, or the complete budget.

\luna{} was acquired earlier than the current reader experiments,
and its historical refusal retries remain in the selected records.
Usage-based costs describe selected calls and exclude earlier
retry attempts; zero recorded usage on blocked calls is not
verified zero billing. \jev{} costs are replay estimates, not
observed online cancellation savings. Legacy Luna logs do not
retain full question text. Their question provenance follows
the collection script's source traces. API model identifiers
do not establish immutable backend revisions.

\section{Per-pair curves}
\label{app:grid}
\begin{figure*}[t]
\centering
\includegraphics[width=\textwidth]{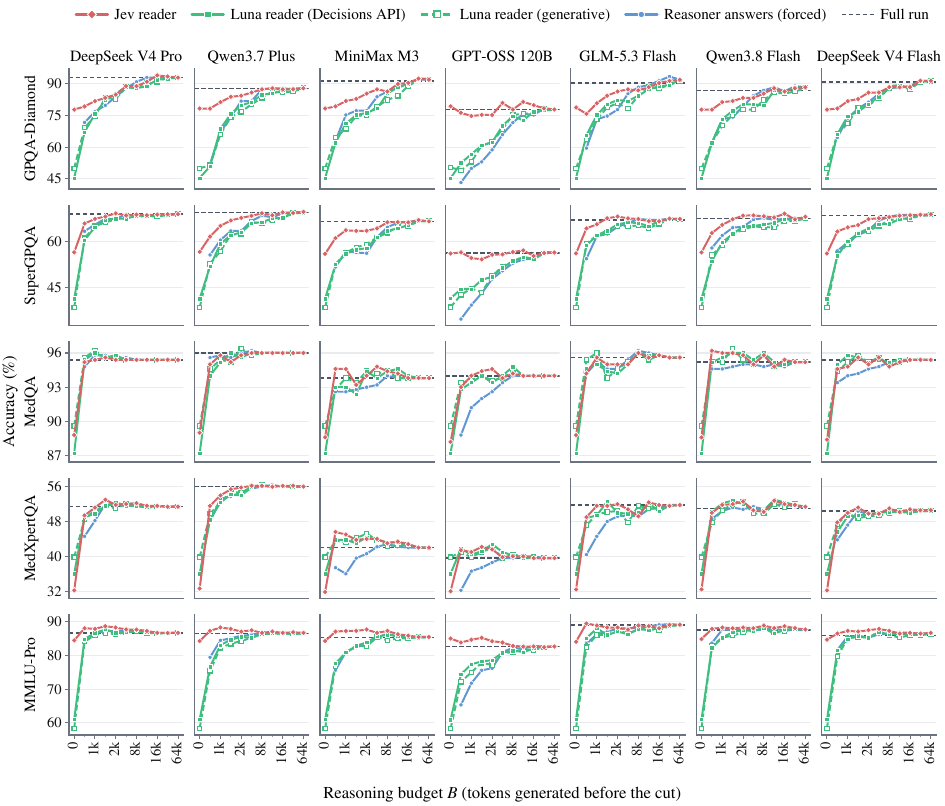}
\caption{Accuracy of the four answerers for every reasoner--benchmark pair against the reasoning budget $B$.
Dashed line: full run.}
\label{fig:grid-budget}
\end{figure*}

\begin{figure*}[t]
\centering
\includegraphics[width=\textwidth]{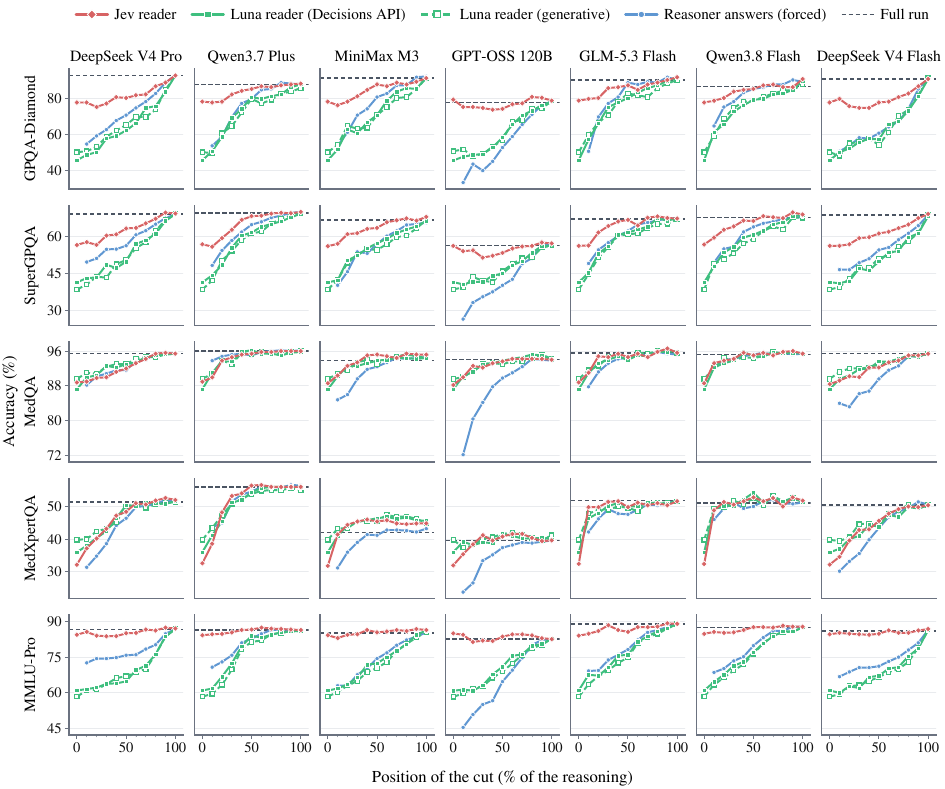}
\caption{As Figure~\ref{fig:grid-budget}, with every question cut at the same fraction of its own reasoning
(diagnostic view).}
\label{fig:grid-fraction}
\end{figure*}

Figures~\ref{fig:grid-budget} and \ref{fig:grid-fraction} report accuracy for all 35 pairs under fixed budgets and relative-progress cuts, respectively.


\section{Stop rules in detail}
\label{app:stop}

\paragraph{Adaptive replay rules.}
\label{app:replayrules}
At checkpoint $j$, \jev{} returns option probabilities $p_j(a)$, with
\begin{equation}
\hat a_j=\arg\max_{a\in\mathcal O}p_j(a),\qquad h_j=\max_{a\in\mathcal O}p_j(a).
\end{equation}
The \emph{conf+stable} rule is
\begin{equation}
j\ge2,\qquad h_j\ge\tau,\qquad \hat a_j=\hat a_{j-1}.
\label{eq:confstable}
\end{equation}
Only the current confidence is thresholded.\\
The \emph{settled+useful} rule is
\begin{equation}
s_j\ge t_s\quad\text{and}\quad u_j\le t_u,
\label{eq:settleduseful}
\end{equation}
where $s_j$ and $u_j$ are separately elicited scores for a submittable answer and the usefulness of further reasoning.
They are returned in the same request and are not derived from option probabilities.
The \emph{either} rule is their logical OR.
The check horizon is 16k tokens, independently of the cap $B$.
Table~\ref{tab:adaptive-replay-procedure} describes the logical replay branches. Replay check overhead is summarized in Appendix~\ref{app:deploy}.

\begin{table}[t]
\centering
\small
\begin{tabular}{@{}r@{\quad}p{0.87\columnwidth}@{}}
\toprule
\multicolumn{2}{@{}l@{}}{\textbf{Procedure: adaptive \jev{} replay}} \\
\midrule
& \textbf{Input:} retained trace, cap $B$, window $w$, interval $k$, selected stopping rule. \\
1 & \textbf{for each} $c\in\{k,2k,\ldots,16{,}000\}$ with $c<\min(B,L)$ \textbf{do} \\
2 & \quad Retrieve \jev{}'s scores from $(q,\mathcal O,\operatorname{Tail}_w(z_{\le c}))$ and charge the check. \\
3 & \quad \textbf{if} the stopping rule triggers \textbf{then} \\
4 & \qquad Submit the current top option and terminate replay. \\
5 & \quad \textbf{end if} \\
6 & \textbf{end for} \\
7 & \textbf{if} generation failed and $L\le B$ \textbf{then} \\
8 & \quad Record an incorrect outcome. Retain generation and check costs, with no final answer call. \\
9 & \textbf{else if} a valid natural answer exists and $L\le B$ \textbf{then} \\
10 & \quad Retain the natural answer. \\
11 & \textbf{else} \\
12 & \quad Retrieve \jev{}'s answer at $\min(B,L)$ and charge the final call. \\
13 & \textbf{end if} \\
\bottomrule
\end{tabular}
\caption{Logical adaptive replay. Earlier checks remain charged when a run later finishes. No additional forced-answer call follows a rule-triggered stop.}
\label{tab:adaptive-replay-procedure}
\end{table}

\paragraph{Policy selection.}
Tuned caps and stopping policies are selected separately for each reasoner--benchmark pair using five-fold cross-validation repeated five times, split by question.
Selection minimizes training cost subject to at most one percentage point of accuracy loss against the full run.
If no candidate is feasible, the candidate with highest training accuracy is selected.
The \emph{auto} setting also chooses the rule family within training folds.
Each question is evaluated using a policy selected using data from the other folds. Its held-out outcomes are then averaged across the five repeats.
The training constraint does not guarantee a corresponding loss bound on held-out questions.
The 1.5k presentation point does not restrict the caps of other policies.
Question-only \jev{} ($B=0$) is a standalone-reader control. Expanded cap grids and cascades appear in Appendix~\ref{app:necessity}.

\paragraph{Selection grids.}
The budget grid is given by $B\in\{0.5,1,1.5,2,4,8,12,16,32,64\}$k.\\
For conf+stable, \\ $\tau\in\{0.5,0.6,0.7,0.8,0.85,0.9,0.95,0.98\}$.\\
For settled+useful, \\ $t_s\in\{0.2,0.3,0.4, 0.5, 0.6, 0.7, 0.8,0.9\}$, and $t_u\in\{0.5,0.6,0.7,0.8,1.0\}$. Either uses the same $t_s,t_u$ grids with $\tau\in\{0.8,0.9,0.95\}$.
Adaptive checks are at 500-token multiples through 16k, strictly before $\min(B,L)$.

Table~\ref{tab:stop-bench} summarizes benchmark-level accuracy and cost savings for the tuned cap and \emph{auto} policies.
\begin{table}[h]
\centering\small
\setlength{\tabcolsep}{3pt}
\begin{tabular}{l cc cc cc}
\toprule
 & \multicolumn{2}{c}{Tuned cap} & \multicolumn{2}{c}{auto} & \multicolumn{2}{c}{Uncapped} \\
\cmidrule(lr){2-3}\cmidrule(lr){4-5}\cmidrule(lr){6-7}
Benchmark & Save & Acc & Save & Acc & & Acc \\
\midrule
GPQA-D     & 38.6 & 87.5 & 59.6 & 85.7 & & 88.3 \\
SuperGPQA  & 64.4 & 65.4 & 77.0 & 65.3 & & 66.3 \\
MedQA      & 57.4 & 94.8 & 60.8 & 94.8 & & 95.1 \\
MedXpertQA & 76.5 & 49.3 & 79.6 & 48.9 & & 48.9 \\
MMLU-Pro   & 79.6 & 87.1 & 79.6 & 87.1 & & 86.1 \\
\bottomrule
\end{tabular}
\caption{Tuned cap and \emph{auto} per benchmark, mean of seven reasoners, held-out folds. Saving in \%
of the full-run cost, accuracy in \%.}
\label{tab:stop-bench}
\end{table}
The range of the per-pair saving difference against the tuned cap is $-8.4$ to $+37.9$ points for
conf+stable, $-6.5$ to $+30.5$ for settled+useful, $-0.4$ to $+38.6$ for either and $0.0$ to $+39.1$ for auto.
On MMLU-Pro, all seven reasoners select a 500-token cap in every fold. No scheduled check precedes this cap, so the stopping policies coincide with the tuned cap.
Table~\ref{tab:picks} lists the most frequent policy selections for each pair.

\begin{table*}[t]
\centering\small
\setlength{\tabcolsep}{4pt}
\begin{tabular}{ll l l r}
\toprule
Reasoner & Benchmark & Tuned cap & \emph{auto}: family (thresholds), cap & $\Delta$Saving \\
\midrule
DeepSeek V4 Pro & GPQA-D & 16k (20) & either (0.95, 0.9, 0.5), 16k (12) & $+13.3$ \\
 & SuperGPQA & 1.5k (14) & s+u (0.2, 1.0), 1k (9) & $+8.5$ \\
 & MedQA & 0.5k (25) & cap, 0.5k (25) & $+0.0$ \\
 & MedXpertQA & 1k (23) & s+u (0.3, 0.7), 1k (13) & $+3.0$ \\
 & MMLU-Pro & 0.5k (25) & cap, 0.5k (25) & $+0.0$ \\
Qwen3.7 Plus & GPQA-D & 8k (19) & either (0.9, 0.9, 0.5), 8k (3) & $+26.1$ \\
 & SuperGPQA & 8k (13) & c+s $\tau{=}$0.5, 4k (14) & $+21.7$ \\
 & MedQA & 0.5k (15) & cap, 0.5k (15) & $+11.2$ \\
 & MedXpertQA & 1.5k (20) & either (0.8, 0.4, 0.8), 1.5k (5) & $+16.3$ \\
 & MMLU-Pro & 0.5k (25) & cap, 0.5k (25) & $+0.0$ \\
MiniMax M3 & GPQA-D & 32k (14) & either (0.95, 0.2, 0.5), 12k (5) & $+39.1$ \\
 & SuperGPQA & 8k (24) & c+s $\tau{=}$0.5, 4k (18) & $+37.9$ \\
 & MedQA & 0.5k (25) & cap, 0.5k (25) & $+0.0$ \\
 & MedXpertQA & 0.5k (25) & cap, 0.5k (25) & $+0.0$ \\
 & MMLU-Pro & 0.5k (25) & cap, 0.5k (25) & $+0.0$ \\
GPT-OSS 120B & GPQA-D & 4k (14) & cap, 0.5k (10) & $+7.6$ \\
 & SuperGPQA & 0.5k (21) & cap, 0.5k (21) & $+3.5$ \\
 & MedQA & 0.5k (14) & cap, 0.5k (14) & $+6.4$ \\
 & MedXpertQA & 0.5k (25) & cap, 0.5k (25) & $+0.0$ \\
 & MMLU-Pro & 0.5k (25) & cap, 0.5k (25) & $+0.0$ \\
GLM-5.3 Flash & GPQA-D & 16k (23) & c+s $\tau{=}$0.85, 16k (8) & $+26.0$ \\
 & SuperGPQA & 1.5k (17) & either (0.8, 0.2, 1.0), 1.5k (14) & $+4.7$ \\
 & MedQA & 1k (19) & s+u (0.2, 1.0), 1k (15) & $+5.9$ \\
 & MedXpertQA & 1k (25) & s+u (0.2, 0.8), 1k (22) & $+2.0$ \\
 & MMLU-Pro & 0.5k (25) & cap, 0.5k (25) & $+0.0$ \\
Qwen3.8 Flash & GPQA-D & 12k (16) & c+s $\tau{=}$0.7, 8k (11) & $+27.9$ \\
 & SuperGPQA & 1.5k (22) & either (0.8, 0.2, 1.0), 1.5k (10) & $+2.9$ \\
 & MedQA & 0.5k (25) & cap, 0.5k (25) & $+0.0$ \\
 & MedXpertQA & 0.5k (14) & cap, 0.5k (14) & $+0.6$ \\
 & MMLU-Pro & 0.5k (25) & cap, 0.5k (25) & $+0.0$ \\
DeepSeek V4 Flash & GPQA-D & 32k (24) & cap, 32k (14) & $+6.9$ \\
 & SuperGPQA & 8k (12) & cap, 2k (7) & $+8.9$ \\
 & MedQA & 0.5k (19) & cap, 0.5k (19) & $+0.0$ \\
 & MedXpertQA & 1k (21) & cap, 1k (21) & $+0.1$ \\
 & MMLU-Pro & 0.5k (25) & cap, 0.5k (25) & $+0.0$ \\
\bottomrule
\end{tabular}
\caption{Most frequent selection over the 25 training folds (count in brackets). c+s: conf+stable ($\tau$);
s+u: settled+useful ($t_s$, $t_u$); either: ($\tau$, $t_s$, $t_u$). $\Delta$Saving: \emph{auto} minus tuned cap,
points of the full run's cost.}
\label{tab:picks}
\end{table*}
When checks occur every 1{,}000 rather than 500 tokens, conf+stable achieves 68.2\% cost saving with an accuracy change of $-0.35$ points. Its saving exceeds that of the tuned cap by 4.9 points, with significantly lower costs in 10 pairs and higher costs in seven. For settled+useful, the corresponding saving is 68.6\% and the accuracy change is $-0.46$ points. Its saving exceeds that of the tuned cap by 5.3 points, with significantly lower costs in 14 pairs and higher costs in one.

\paragraph{GPQA same-budget diagnostic.}
\label{app:gpqaaudit}
An offline recomputation on 2026-10-11 reproduces all seven GPQA cells in the current
\texttt{stop\_deploy\_a9} snapshot. Mean full accuracy is 88.32\%; tuned cap gives 87.49\% at 38.56\%
cost saving, and \emph{auto} gives 85.73\% at 59.55\% saving. Its additional 1.76pp loss against cap
combines different selected budgets and within-budget stopping. Disabling the rule while retaining
each fold's selected \emph{auto} budget yields 86.97\% accuracy: the same-budget rule difference is
$-1.24$pp, and the remaining budget-selection difference is $-0.52$pp. This is a post-hoc diagnostic, not an independently fitted deployment policy.
All 175 \emph{auto} training folds have a candidate satisfying the one-point training constraint, but 95 test-fold point estimates exceed a one-point loss. The overlapping folds are not independent trials.
After averaging the five CV repeats per question, a paired bootstrap jointly resampling original
question IDs across the seven reasoners (10{,}000 draws) gives a 95\% interval of $[-4.49,-0.73]$pp
for \emph{auto} versus full and $[-2.33,-0.23]$pp for enabling the rule at the same budgets.
These intervals condition on the recorded selections, do not refit CV and are not multiplicity-adjusted.
The interval for \emph{auto} versus full includes a one-point loss. Although the estimated loss exceeds the selection target, the interval does not establish that the population loss exceeds that target.

\section{SuperGPQA by difficulty}
\label{app:difficulty}
Table~\ref{tab:difficulty} reports fixed-budget accuracy and relative cost by official difficulty level, averaged over the seven reasoners.

\begin{table}[h]
\centering\small
\setlength{\tabcolsep}{3pt}
\begin{tabular}{l ccc}
\toprule
 & Easy & Middle & Hard \\
\midrule
Full run           & 69.1 & 67.7 & 61.0 \\
Forced, 1.5k       & 67.5 (52.5) & 63.1 (36.6) & 45.4 (19.0) \\
\jev{}, 1.5k       & 69.2 (48.5) & 66.9 (32.7) & 56.1 (16.3) \\
\luna{}-dec., 1.5k & 68.4 (49.7) & 64.4 (34.0) & 44.8 (17.2) \\
\luna{}-gen., 1.5k & 67.2 (49.5) & 64.2 (33.7) & 43.7 (17.0) \\
\bottomrule
\end{tabular}
\caption{SuperGPQA difficulty, accuracy in \% (cost as \% of the full run). Questions are unweighted within each level, then seven reasoners are macro-averaged. The original sample contains 167 / 166 / 167 questions. One Qwen3.8 Flash outcome is excluded from the matched 1.5k comparison because its exact forced answer is unavailable. Tuned-policy results under the main selection protocol are reported in Table~\ref{tab:stop-bench}.}
\label{tab:difficulty}
\end{table}
At 1.5k tokens, \jev{} reaches 69.2\%, 66.9\% and 56.1\% on easy, middle and hard questions, respectively, compared with 69.1\%, 67.7\% and 61.0\% for the full run.
The \jev{} advantage is concentrated on hard questions, where cuts are most frequent; 157 of the 167 hard
questions are science or engineering.
This analysis was added after the main results and is exploratory.

\section{Completed reasoning per pair}
\label{app:complete}
Table~\ref{tab:complete} compares the reasoner's natural answer with \jev{} reading completed reasoning, restricted to runs with valid final answers.

\begin{table*}[t]
\centering\small
\setlength{\tabcolsep}{4pt}
\begin{tabular}{ll r cc c c c c}
\toprule
Benchmark & Model & $n$ & R Acc. (\%) & \jev{} Acc. (\%) & $\Delta$ (pp) & 95\% CI (pp) & Agree (\%) & \jev{}+ / R+ \\
\midrule
GPQA-D & DeepSeek V4 Pro & 197 & 93.4 & 93.4 & +0.0 & [0.0, 0.0] & 100.0 & 0 / 0 \\
 & Qwen3.7 Plus & 198 & 87.9 & 88.4 & +0.5 & [0.0, 1.5] & 98.5 & 1 / 0 \\
 & MiniMax M3 & 194 & 93.3 & 93.3 & +0.0 & [-1.5, 1.5] & 99.0 & 1 / 1 \\
 & GPT-OSS 120B & 194 & 77.8 & 78.9 & +1.0 & [0.0, 2.6] & 98.5 & 2 / 0 \\
 & GLM-5.3 Flash & 190 & 94.2 & 94.2 & +0.0 & [0.0, 0.0] & 100.0 & 0 / 0 \\
 & Qwen3.8 Flash & 189 & 91.0 & 93.1 & +2.1\rlap{\textsuperscript{*}} & [0.5, 4.2] & 97.9 & 4 / 0 \\
 & DeepSeek V4 Flash & 195 & 92.3 & 92.3 & +0.0 & [0.0, 0.0] & 99.5 & 0 / 0 \\
\midrule
SuperGPQA & DeepSeek V4 Pro & 499 & 69.0 & 69.3 & +0.3 & [-0.3, 1.1] & 99.4 & 2 / 1 \\
 & Qwen3.7 Plus & 499 & 69.5 & 70.0 & +0.5 & [-0.2, 1.4] & 99.0 & 4 / 1 \\
 & MiniMax M3 & 496 & 67.0 & 68.3 & +1.2\rlap{\textsuperscript{*}} & [0.3, 2.3] & 98.0 & 6 / 0 \\
 & GPT-OSS 120B & 494 & 56.4 & 57.1 & +0.7\rlap{\textsuperscript{*}} & [0.2, 1.4] & 98.0 & 4 / 0 \\
 & GLM-5.3 Flash & 493 & 67.8 & 67.6 & -0.2 & [-0.5, 0.0] & 99.6 & 0 / 1 \\
 & Qwen3.8 Flash & 493 & 68.2 & 69.2 & +1.0\rlap{\textsuperscript{*}} & [0.3, 1.8] & 97.8 & 6 / 0 \\
 & DeepSeek V4 Flash & 495 & 69.1 & 69.2 & +0.1 & [-0.9, 0.9] & 98.8 & 3 / 2 \\
\midrule
MedQA & DeepSeek V4 Pro & 500 & 95.4 & 95.4 & +0.0 & [0.0, 0.0] & 100.0 & 0 / 0 \\
 & Qwen3.7 Plus & 500 & 96.0 & 96.0 & +0.0 & [0.0, 0.0] & 100.0 & 0 / 0 \\
 & MiniMax M3 & 499 & 94.0 & 95.4 & +1.4\rlap{\textsuperscript{*}} & [0.4, 2.4] & 98.6 & 7 / 0 \\
 & GPT-OSS 120B & 500 & 94.0 & 94.0 & +0.0 & [0.0, 0.0] & 100.0 & 0 / 0 \\
 & GLM-5.3 Flash & 500 & 95.6 & 95.6 & +0.0 & [0.0, 0.0] & 100.0 & 0 / 0 \\
 & Qwen3.8 Flash & 500 & 95.2 & 95.4 & +0.2 & [0.0, 0.6] & 99.8 & 1 / 0 \\
 & DeepSeek V4 Flash & 499 & 95.6 & 95.6 & +0.0 & [0.0, 0.0] & 100.0 & 0 / 0 \\
\midrule
MedXpertQA & DeepSeek V4 Pro & 500 & 51.4 & 52.0 & +0.6 & [0.0, 1.4] & 99.4 & 3 / 0 \\
 & Qwen3.7 Plus & 500 & 56.0 & 56.0 & +0.0 & [0.0, 0.0] & 99.8 & 0 / 0 \\
 & MiniMax M3 & 499 & 42.1 & 44.9 & +2.8\rlap{\textsuperscript{*}} & [1.0, 4.6] & 94.2 & 17 / 3 \\
 & GPT-OSS 120B & 500 & 39.6 & 39.6 & +0.0 & [0.0, 0.0] & 99.6 & 0 / 0 \\
 & GLM-5.3 Flash & 500 & 51.8 & 51.6 & -0.2 & [-0.6, 0.0] & 99.6 & 0 / 1 \\
 & Qwen3.8 Flash & 496 & 51.4 & 51.8 & +0.4 & [-0.4, 1.2] & 98.4 & 3 / 1 \\
 & DeepSeek V4 Flash & 497 & 50.7 & 50.7 & +0.0 & [-0.8, 0.8] & 99.2 & 2 / 2 \\
\midrule
MMLU-Pro & DeepSeek V4 Pro & 500 & 86.6 & 87.0 & +0.4 & [0.0, 1.0] & 99.6 & 2 / 0 \\
 & Qwen3.7 Plus & 500 & 86.4 & 86.4 & +0.0 & [0.0, 0.0] & 100.0 & 0 / 0 \\
 & MiniMax M3 & 499 & 85.4 & 86.4 & +1.0\rlap{\textsuperscript{*}} & [0.2, 2.0] & 98.8 & 5 / 0 \\
 & GPT-OSS 120B & 499 & 82.6 & 82.6 & +0.0 & [-0.6, 0.6] & 99.2 & 1 / 1 \\
 & GLM-5.3 Flash & 500 & 89.0 & 89.0 & +0.0 & [0.0, 0.0] & 99.6 & 0 / 0 \\
 & Qwen3.8 Flash & 500 & 87.4 & 87.8 & +0.4 & [0.0, 1.0] & 99.4 & 2 / 0 \\
 & DeepSeek V4 Flash & 495 & 86.7 & 86.9 & +0.2 & [0.0, 0.6] & 99.8 & 1 / 0 \\
\bottomrule
\end{tabular}
\caption{Completed reasoning: the reasoner's own answer versus \jev{} reading the final approximately 1{,}000 reasoning tokens, on 15{,}309 runs with a valid natural answer. Agree is exact option-letter agreement. The last column counts outcomes where only \jev{} / only the reasoner is correct (77 / 14 in total). $^{*}$ marks an uncorrected 95\% paired-bootstrap interval excluding zero (2{,}000 resamples).}
\label{tab:complete}
\end{table*}

\section{Model--label disagreement}
\label{app:noise}
Table~\ref{tab:noise} reports accuracy gaps before and after removing flagged questions.

\begin{table}[h]
\centering\small
\setlength{\tabcolsep}{3.5pt}
\resizebox{\columnwidth}{!}{%
\begin{tabular}{l c cc cc}
\toprule
 & & \multicolumn{2}{c}{\jev{} $-$ forced} & \multicolumn{2}{c}{\jev{} $-$ \luna{} dec.} \\
\cmidrule(lr){3-4}\cmidrule(lr){5-6}
Benchmark & Flagged & All & Removed & All & Removed \\
\midrule
GPQA-D     & 4 & 9.2 & 9.4 & 6.5 & 7.3 \\
SuperGPQA  & 49 & 5.0 & 5.7 & 4.3 & 4.9 \\
MedQA      & 1 & 0.9 & 0.9 & 0.1 & 0.1 \\
MedXpertQA & 66 & 2.4 & 2.9 & 0.6 & 0.7 \\
MMLU-Pro   & 20 & 3.1 & 3.3 & 2.9 & 3.0 \\
\bottomrule
\end{tabular}}
\caption{Accuracy gaps (points) at $B=1.5$k with all questions and with flagged questions removed, mean of
seven reasoners.}
\label{tab:noise}
\end{table}
Agreement of all seven runs on a non-gold option warrants label review but is not evidence of a label error:
models may share knowledge gaps or misinterpret a question similarly.
GPQA-Diamond items 106 and 17 meet this agreement criterion.
Of items 17, 89, and 126 checked by hand, only item 17 has independent support for a wrong label; item 89 is ambiguous and item 126 repeats a distractor.
Flagged counts are therefore not label-error rates, and findings on MMLU \citep{gema2025mmlu} do not establish errors in our MMLU-Pro items.
At $B=1.5$k, removing the 140 flagged questions changes the mean \jev{}--forced gap from $+4.1$pp to $+4.4$pp. In both cases, uncorrected paired-bootstrap intervals identify higher accuracy in 21 pairs and lower accuracy in none. 


\section{Is reasoning needed? Question-only \jev{} as a policy}
\label{app:necessity}
We add two kinds of policy to the out-of-fold policy selection.
(i) $B=0$ in the cap grid: \jev{} answers from the question alone and the reasoner is not called.
(ii) Cascades: \jev{} first answers from the question alone. If its top probability is at least $g$, the answer is submitted. Otherwise, the question is escalated either to the full run (\emph{\jev{}$\to$full}) or to a cap with \jev{} reading the prefix (\emph{\jev{}$\to$cap}).
The gate threshold $g$ (0.3 to 0.98, or never) and $B$ are selected inside the training folds; the gate call is charged to every question.
Because a single tolerance hides the trade-off, Table~\ref{tab:necessity} reports each policy at four training-fold tolerances.

\begin{table}[!h]
\centering\small
\setlength{\tabcolsep}{3pt}
\resizebox{\columnwidth}{!}{%
\begin{tabular}{c cccc}
\toprule
Tol. & Cap & Cap with $B{=}0$ & \jev{}$\to$full & \jev{}$\to$cap \\
\midrule
0.25 & 57.6 / $+0.05$ & 59.1 / $+0.07$ & 33.0 / $-0.29$ & 71.4 / $-0.70$ \\
0.5  & 59.7 / $+0.03$ & 61.4 / $+0.04$ & 35.1 / $-0.40$ & 73.3 / $-0.83$ \\
1    & 63.3 / $-0.13$ & 65.2 / $-0.17$ & 39.6 / $-0.79$ & 76.9 / $-1.21$ \\
2    & 68.7 / $-0.83$ & 71.5 / $-0.96$ & 48.2 / $-1.50$ & 82.2 / $-2.17$ \\
\bottomrule
\end{tabular}}
\caption{Cost saving (\%) / accuracy change (points) against the full run on held-out folds, mean over 35 pairs.
Tol.: accuracy loss allowed on the training folds. References: fixed 1.5k with \jev{} 65.5 / $-1.13$;
\jev{} alone 98.3 / $-8.97$; stop rule \emph{auto} (tol.\ 1) 71.3 / $-0.58$.}
\label{tab:necessity}
\end{table}

\paragraph{Selecting the question-only baseline.}
At tolerance 1, allowing $B=0$ selects question-only \jev{}
in most folds for four pairs: GPT-OSS 120B on GPQA-Diamond, SuperGPQA and MMLU-Pro, and MiniMax M3 on MMLU-Pro. Across 35 pairs, mean cost saving increases from 63.3\% to 65.2\%, with a mean accuracy change of $-0.03$ points relative to the tuned cap.
Costs are significantly lower in eight pairs and higher in none, whereas accuracy is significantly lower in four pairs (unadjusted 95\% paired-bootstrap intervals).

\paragraph{Escalating to the full run is not competitive.}
Answering confident questions with \jev{} alone and running the reasoner in full on the rest saves only 33--48\%,
less than any cap at a similar accuracy: buying the whole trace for uncertain questions costs more than buying a short prefix for all of them.

\paragraph{Escalating to a cap is.}
\jev{}$\to$cap lies on a better frontier than the cap alone: at $-0.83$pp it saves 73.3\% against 68.7\%,
and at about 71.5\% saving it loses 0.70pp against 0.96pp.
It is comparable to the \emph{auto} stop rule.
The gate submits 79--93\% of MedQA and MMLU-Pro questions, 53\% of GPQA-Diamond and about 21\% of MedXpertQA.
On held-out folds, overall accuracy drops by 1.21pp relative to the full run, exceeding the 1pp tolerance
used for policy selection on training folds.

\section{Exact cut positions for the forced answer}
\label{app:exact}
Table~\ref{tab:exact} reports current exact-position coverage. The 1.5k comparison uses the completed exact-forced overlay, and other budgets retain their existing forced records and earlier-position fallback.

\begin{table}[h]
\centering\small
\setlength{\tabcolsep}{3pt}
\resizebox{\columnwidth}{!}{%
\begin{tabular}{r cc cc cc}
\toprule
 & Exact & Offset & \multicolumn{2}{c}{\jev{} $-$ forced} & \multicolumn{2}{c}{Sig.\ better / worse} \\
\cmidrule(lr){4-5}\cmidrule(lr){6-7}
$B$ & (\%) & (tok.) & All cut & Exact only & All cut & Exact only \\
\midrule
0.5k & 93.7 & 500 & $+11.5$ & $+11.3$ & 29 / 0 & 30 / 0 \\
1k   & 92.7 & 668 & $+8.8$ & $+8.0$ & 25 / 0 & 21 / 0 \\
1.5k & 100.0 & -- & $+8.6$ & $+8.6$ & 21 / 0 & 21 / 0 \\
2k   & 88.4 & 631 & $+8.1$ & $+7.6$ & 18 / 0 & 18 / 0 \\
4k   & 82.1 & 615 & $+5.3$ & $+4.5$ & 10 / 0 & 8 / 0 \\
8k   & 64.9 & 600 & $+3.6$ & $+2.7$ & 3 / 0 & 3 / 1 \\
16k  & 59.4 & 826 & $+3.1$ & $+4.2$ & 4 / 0 & 4 / 0 \\
\bottomrule
\end{tabular}}
\caption{Exact coverage is averaged over pairs. Offset is the mean fallback distance in nominal tokens. Accuracy gaps condition on cut outcomes using the main analysis's question weights. The exact-only subset requires at least five cut outcomes per pair (32 eligible pairs at 8k; 25 at 16k). Significance uses 100{,}000 paired-bootstrap resamples, uncorrected. At 1.5k, all 6{,}500 matched cut outcomes have exact forced answers; no fallback offset applies.}
\vspace{-2mm}
\label{tab:exact}
\end{table}

\section{Forced-answer protocol}
\label{app:protocol}
GPT-OSS 120B and GLM-5.3 Flash use think-close; the other five reasoners use strict forced answering. At $B=1.5$k, the mean \jev{}--forced gap is 2.9pp over the 25 strict pairs and 7.2pp over the 10 think-close pairs. Uncorrected paired-bootstrap intervals identify higher accuracy for \jev{} in 13 strict pairs and eight think-close pairs, with no lower-accuracy pairs. On actual cuts, the respective gaps are 6.9pp and 12.8pp. This protocol grouping is descriptive and confounded with reasoner identity.

\section{Question-only baselines and trace contribution}
\label{app:priorcontrol}
Tables~\ref{tab:prior} and \ref{tab:priorcontrol} report trace gains and accuracy conditioned on question-only correctness, respectively.

\begin{table}[h]
\centering\small
\setlength{\tabcolsep}{3.5pt}
\begin{tabular}{l c ccc cc}
\toprule
 & & \multicolumn{3}{c}{Net change (pts)} & \multicolumn{2}{c}{At 100\% (\%)} \\
\cmidrule(lr){3-5}\cmidrule(lr){6-7}
Reader & Q-only & 10\% & 50\% & 100\% & Fixed & Hurt \\
\midrule
\jev{}            & 66.7 & $+2.5$ & $+7.0$  & $+9.2$  & 43.1 & 7.7 \\
\luna{} dec.      & 56.0 & $+3.5$ & $+12.4$ & $+19.8$ & 56.1 & 8.8 \\
\luna{} gen.      & 56.5 & $+2.8$ & $+11.3$ & $+19.1$ & 55.6 & 9.1 \\
\bottomrule
\end{tabular}
\caption{Question-only accuracy (Q-only) and net change at the given trace fraction, pooled over 15{,}030 model--question outcomes from 2{,}182 unique questions (unweighted, including SuperGPQA).
Fixed: share of incorrect question-only answers corrected; hurt: share of correct question-only answers made incorrect.}
\label{tab:prior}
\end{table}

\begin{table}[h]
\centering\small
\setlength{\tabcolsep}{3.5pt}
\resizebox{\columnwidth}{!}{%
\begin{tabular}{l r ccc}
\toprule
Subset & Cut at & \jev{} & \luna{} dec. & \luna{} gen. \\
\midrule
All question-only answers    & 10\%  & 13.4 & 12.8 & 14.5 \\
incorrect ($n=3{,}455$)      & 50\%  & 27.1 & 26.7 & 25.5 \\
                   & 100\% & 34.2 & 34.0 & 33.7 \\
\midrule
All question-only answers   & 10\%  & 97.8 & 96.5 & 94.6 \\
correct ($n=6{,}597$)      & 50\%  & 96.9 & 95.7 & 95.3 \\
                   & 100\% & 96.3 & 96.2 & 96.3 \\
\bottomrule
\end{tabular}}
\caption{Accuracy (\%) after reading a prefix, on outcomes where all three readers are wrong (top) or all right (bottom) from the question alone; pooled over pairs, unweighted. Counts are model--question outcomes, not independent questions.}
\label{tab:priorcontrol}
\end{table}
This complete-case diagnostic requires parseable answers from all readers at 0, 10, 30, 50, 80 and
100\%, excluding 347 of the 15{,}377 retained outcomes. It retains saved reasoning from no-valid-natural-answer runs, including one failed-generation trace. Unlike the primary evaluation, excluded refusals or parsing failures are not counted as wrong in this diagnostic. Selection and weighting differences explain why its question-only accuracy differs from the budget evaluation. Inputs use the final nominal 1k-token window, and 100\% denotes the end of the observed trace.
When all three readers start wrong, their overall recovery rates differ by at most about 1.7pp at the listed cuts. The similar point estimates for complete traces do not establish equivalence.
When all three start right, \jev{} keeps the right answer more often after reading an early prefix ($+1.3$ points
over \luna{} decision at 10\%, $+4.5$ on SuperGPQA, $+3.4$ on MMLU-Pro), and the difference vanishes on the complete
trace.
These controls distinguish recovery from preserving a correct initial answer, but do not identify a
reader's internal knowledge or mechanism. Legacy pooled intervals resample model--question outcomes without accounting for shared question IDs across reasoners. We therefore report these rates descriptively and do not use the intervals to claim superiority or equivalence.

\section{Questions with a usable full-run answer}
\label{app:answered}
Of 15{,}377 retained runs, 67 no-valid-answer rerun outcomes and one saved failed-generation outcome count as incorrect for the full run. They are excluded only from this valid-natural-answer analysis.
On the remaining 15{,}309 runs, the mean full-run accuracy is 77.4\%. At $B=1.5$k, \jev{} reaches 76.0\% ($-1.4$pp versus full), \luna{} decision 73.1\%, \luna{} generative 72.8\%, and forced answering 72.0\% ($-5.4$pp). These are macro-averages over the same 35 pairs.

\section{Multiple comparisons and loss bounds}
\label{app:multiple}
Uncorrected per-pair intervals are exploratory and do not control a single family across all budgets and comparators. For \jev{} versus forced answering at $B=1.5$k, we test each pair's accuracy difference by paired question bootstrap, stratified and weighted by difficulty on SuperGPQA, using $R=100{,}000$ resamples. 

The percentile-tail quantity is
\[
\begin{aligned}
p_{\mathrm{tail}}
&= \min\!\left\{
\widehat{\Pr}(\hat\Delta^*\le0),
\widehat{\Pr}(\hat\Delta^*\ge0)
\right\},\\
p
&= \max\!\left\{
R^{-1},\min(1,2p_{\mathrm{tail}})
\right\}.
\end{aligned}
\]
We apply Holm's step-down correction \citep{holm1979simple} across the 35 pairs at $\alpha=0.05$. It retains 17 higher-accuracy pairs and none lower; these counts are unchanged across seeds 0, 1 and 2. Percentile-tail tests can be poorly calibrated with few discordant outcomes. We therefore also report exact paired tests with the same correction: exact McNemar tests for the four unweighted benchmarks and an exact weighted answerer-label-swap test for SuperGPQA. The latter assumes within-question answerer-label exchangeability, a stronger null than zero weighted mean difference. This sensitivity check retains 14 higher-accuracy pairs and none lower.

At $B=1.5$k, the macro-average \jev{}--full change is $-1.13$pp with a one-sided 95\% lower bound of $-1.74$pp. On the valid-natural-answer subset the corresponding values are $-1.39$pp and $-1.98$pp. Macro bounds use 20{,}000 paired question-cluster resamples, stratified by SuperGPQA difficulty and sharing question multiplicities across reasoners. Per-pair bounds use paired question resampling; only 12 of 35 pairs have a lower bound above $-1$pp. Nonsignificance does not establish equivalence or a universal one-point loss guarantee.

\section{Replay overhead of stopping rules}
\label{app:deploy}
Table~\ref{tab:deploy} reports replayed calls per question, stopping outcomes, and cost savings for each policy.

The \emph{auto} policy makes 1.5 \jev{} calls per question on average, 4.0 on GPQA-Diamond and about 0.5 on MMLU-Pro and MedQA.
The \emph{conf+stable} policy is more expensive than the cap in nine pairs, mostly on the medical sets and for DeepSeek V4 Flash (GPQA$-$Diamond: $-8.4$~pp), where each check is a larger share of a cheap run. In the observed held-out results, auto is no costlier than the tuned cap in any pair.

\begin{table}[!t]
\centering\small
\setlength{\tabcolsep}{3pt}
\resizebox{\columnwidth}{!}{%
\begin{tabular}{l c c ccc c}
\toprule
 & Saving & \jev{} calls & \multicolumn{3}{c}{Questions (\%)} & Costlier \\
\cmidrule(lr){4-6}
Policy & (\%) & per question & Rule & Cap & Finished & than cap \\
\midrule
Tuned cap      & 63.3 & 0.48 & --   & 48.2 & 51.8 & -- \\
conf+stable    & 68.8 & 1.73 & 20.0 & 40.9 & 39.1 & 9 (9) \\
settled+useful & 68.7 & 1.92 & 28.7 & 39.0 & 32.3 & 4 (1) \\
either         & 70.8 & 1.69 & 30.1 & 38.0 & 31.9 & 1 (0) \\
auto           & 71.3 & 1.52 & 27.3 & 37.9 & 34.7 & 0 (0) \\
\bottomrule
\end{tabular}}
\caption{Stop rules on held-out folds, mean over 35 pairs. Rule / Cap / Finished: share of questions stopped by the rule, cut at the cap, or finished within the cap. Costlier than cap: pairs where the rule costs more than the tuned cap (significantly in brackets).}
\label{tab:deploy}
\end{table}

\end{document}